\documentclass[journal]{IEEEtran}

\usepackage{balance}
\usepackage{times}
\usepackage{epsfig}
\usepackage{graphicx}
\usepackage{amsmath}
\usepackage{amssymb}
\usepackage{multirow}
\usepackage{amsmath}
\usepackage{xcolor}
\usepackage{caption}
\usepackage[linesnumbered,ruled]{algorithm2e}
\usepackage{microtype}

\def\ie{{\em i.e.}}
\def\eg{{\em e.g.}}
\def\etal{{\em et al.}}

\newcommand{\figref}[1]{Fig. \ref{#1}}
\newcommand{\tabref}[1]{Tab. \ref{#1}}
\newcommand{\equref}[1]{(\ref{#1})}
\newcommand{\secref}[1]{Section \ref{#1}}
\newcommand{\algref}[1]{Alg. \ref{#1}}

\newcommand{\mc}[1]{\mathcal{#1}}

\newcommand{\bm}[1]{\mbox{\boldmath{$#1$}}}
\newcommand{\br}[1]{\bm{\mathrm{#1}}}
\newcommand{\bs}[1]{\boldsymbol{\texttt{#1}}}

\usepackage{epstopdf}
\usepackage{ragged2e}
\usepackage{booktabs}
\usepackage{color}
\usepackage{multirow}
\usepackage{bm}
\usepackage{bbm}
\usepackage[misc]{ifsym}
\usepackage{graphicx}
\usepackage{amsmath}
\usepackage{amsfonts}
\usepackage{enumerate}
\usepackage[colorlinks,linkcolor=black]{hyperref}

\begin{document}
\title{DanceIt: Music-inspired Dancing Video Synthesis}

\author{Xin~Guo,~Yifan~Zhao,~Jia~Li,~\IEEEmembership{Senior Member,~IEEE}\\
\IEEEcompsocitemizethanks{\IEEEcompsocthanksitem X. Guo, Y. Zhao, J. Li are with the State Key Laboratory of Virtual Reality Technology and Systems, School of Computer Science and Engineering, Beihang University, Beijing, 100191, China.
\IEEEcompsocthanksitem J. Li is also with the Peng Cheng Laboratory, Shenzhen 518000, China (e-mail:
jiali@buaa.edu.cn)

J. Li is the corresponding author. URL: http://cvteam.net
}}

\markboth{}%
{Guo \MakeLowercase{\textit{et al.}}: Bare Demo of IEEEtran.cls for IEEE Journals}

\maketitle

\begin{figure*}[t]
\begin{center}
   \includegraphics[width=0.9\linewidth]{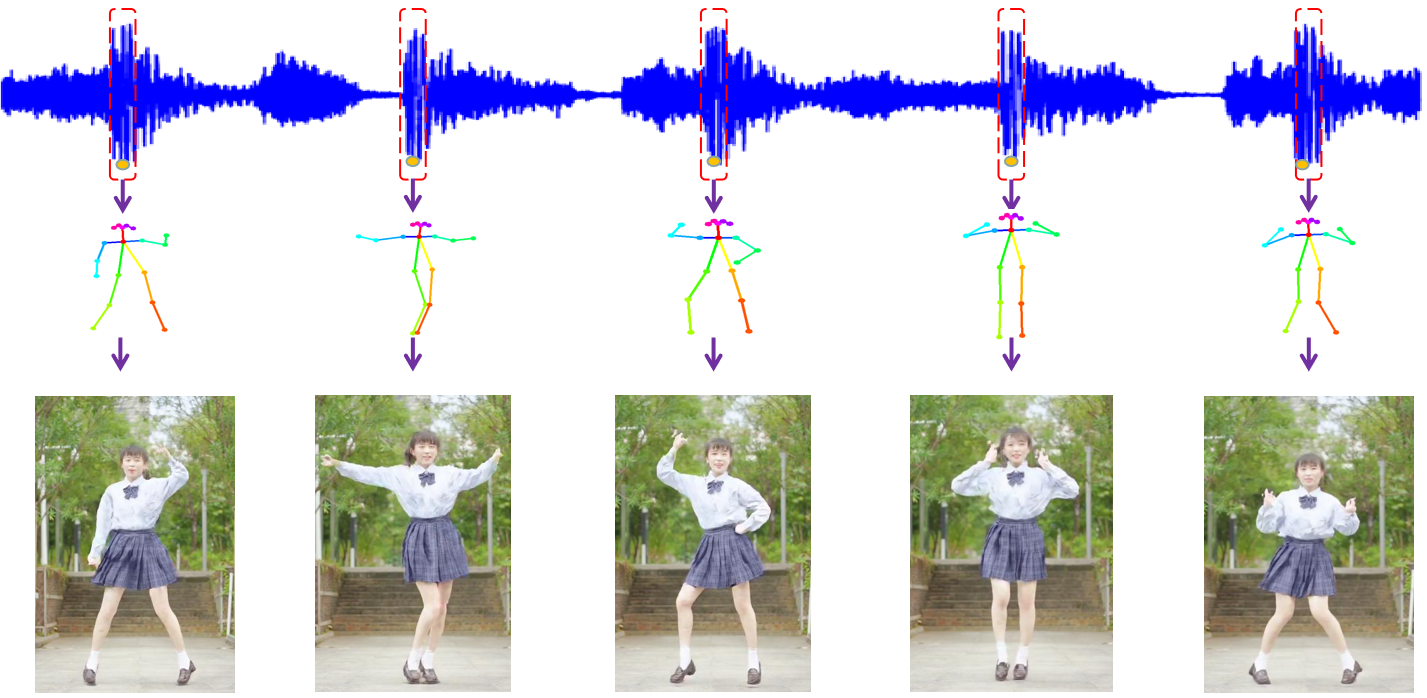}
\end{center}
   \caption{The illustration of our motivation. Given a piece of input audio, our motif is to generate a dancing sequence automatically by taking the human pose as an intermediary. The representative pose and generated dancing frames corresponding to the audio peaks (in the red dotted box) are viewed in the second and third rows. }
\label{fig:motivation}
\end{figure*}
\begin{abstract}
Close your eyes and listen to music, one can easily imagine an actor dancing rhythmically along with the music. These dance movements are usually made up of dance movements you have seen before. In this paper, we propose to reproduce such an inherent capability of the human-being within a computer vision system. The proposed system consists of three modules. To explore the relationship between music and dance movements, we propose a cross-modal alignment module that focuses on dancing video clips, accompanied by pre-designed music, to learn a system that can judge the consistency between the visual features of pose sequences and the acoustic features of music. The learned model is then used in the imagination module to select a pose sequence for the given music. Such pose sequence selected from the music, however, is usually discontinuous. To solve this problem, in the spatial-temporal alignment module we develop a spatial alignment algorithm based on the tendency and periodicity of dance movements to predict dance movements between discontinuous fragments. In addition, the selected pose sequence is often misaligned with the music beat. To solve this problem, we further develop a temporal alignment algorithm to align the rhythm of music and dance. Finally, the processed pose sequence is used to synthesize realistic dancing videos in the imagination module. The generated dancing videos match the content and rhythm of the music. Experimental results and
subjective evaluations show that the proposed approach can perform the function of generating promising dancing videos by inputting music.
\end{abstract}


\IEEEpeerreviewmaketitle

\section{Introduction}\label{sec:intro}

In an imagined room of the human-being, the rhythm of Hip-hop music can be incarnated as a dynamic street dancer, and the music of \emph{Swan Lake} usually leads to an elegant ballet dancer. Such a capability of cross-modal imagination is also a crucial foundation of the so-called creativity that makes human brains different from a computer. With the rapid development of computing devices and machine learning technologies, a concern naturally arises: is it possible to reproduce such an inherent capability of the human-being within a computer vision system?

Keeping this in our mind, the main motif of our paper is to generate the natural dancing sequences along with the musical beats (as shown in~\figref{fig:motivation}), namely audio-inspired video synthesis.
However, the dancing sequences to generate and musical audio are substantially two different modalities of media and generating from one modal to another is not a one-to-one mapping problem. Thus there exist two burning challenges to be solved: 1) how to generate a dancing frame corresponding to the musical beats; 2) how to make the video sequence vivid and natural as human acting?

Actually, many research efforts~\cite{chang2008system,suwajanakorn2017synthesizing,tan2019text2scene,li2019object} have been done to explore the generation task across multiple modalities.
Among these works, the most explored transformation is between text and other media formats. For example, Chang \emph{et al.} \cite{chang2008system} proposed to generate text descriptions from given video clips. Tan \emph{et al.} \cite{tan2019text2scene} and Li \emph{et al.} \cite{li2019object} focused on a reverse problem that intended to synthesize realistic images from texts.

Beyond these efforts in other modalities, there are few works focusing on this less-explored task. Shlizerman \emph{et al.} \cite{shlizerman2018audio} proposed a deep regression model to predict the sequence of playing violin and piano along with the specific musical note. However, this work mainly focuses on the 3D localization of a fixed animation model, which still remains a gap from the video in the wild scenarios.
As the most relevant research of our work, Tang \emph{et al.} \cite{tang2018dance} developed an end-to-end LSTM-autoencoder structure to predict dancing pose sequences corresponding to the music melody.
Lee~\etal~\cite{lee2019dancing} proposed a decomposition-to-composition network to align the dancing skeletons with musical beats.
However, generating visual content with audio in the frequency domain is indeed an ill-posed problem. There may exist enormous possible dancing movements when hearing the same music piece. In addition, these generated sequences also face difficulties in some complex movements,~\eg, backside or turning around. Therefore in this paper, we propose to solve this problem by building correspondences across different modalities rather than learning one-to-one bijections, while the latter one usually leads to over-fitting in network training.


To solve this drawback and the first challenge as well, we resort to pose skeletons as intermediaries to bridge the connection between the audio and visual content (illustrated in~\figref{fig:motivation}). Instead of estimating specific poses aligned with each musical beat, we propose to learn the coefficient relationship between these two modalities. Thus a deep metric learning approach is developed to learn this co-efficiency when adopting \textit{pose fragment} as the basic unit. To further solve the second challenge, we further propose to align different fragments with musical beats and refine the fragment transmissions, making the generated video vivid and natural.

Taken these two cues together, we tentatively propose a unified framework to synthesize the audio-inspired videos, which are leading by two alignment modules corresponding to the two main challenges. Given an input audio clip, the first cross-modal alignment module aims to find the correlations between audio clips and pose fragments. Taken the advantage of MFCC~\cite{zheng2001comparison} and bidirectional LSTM~\cite{zhang2015bidirectional}, the audio clips are first encoded by the audio encoder to transform the audio from the frequency domain into low-dimension features.  And the corresponding pose skeletons are encoded by a graph convolutional network to understand the sparse pose skeletons. Thus the pose and audio features from two different modalities can be embedded into one unified space, and then the correlations between them can be learned by minimizing the feature distances (see ~\figref{fig:method}).

With the learned correlations, in the generation phase, we retrieve the best-matched pose fragment for each audio clip and concatenate them as an integral pose sequence. However, there usually exist sudden changes between two matched pose fragments and the pre-stored pose fragments may not perfectly match the audio beats, which makes the generated video seems unnatural. In order to solve this challenge, we propose a spatial-temporal alignment module that refines the pose sequences from two different perspectives.
In the spatial domain, we first detect the sudden changes in the pose sequences and refine them by a Time Series Decomposition formula (TSD). In the temporal domain, we extract the musical beats (high-frequency points) as key cues and enforce the local maximum pose movements matching with the extracted beats. Hence the generated dancing video can show the notable pose movement (\eg, raising arms) aligning with the musical beats.

After obtaining the natural dancing sequence, we follow the commonly-used pose-to-video generation to make a complete synthesis. Specially, we further adopt~\cite{chan2019everybody} with FaceGAN to generate a clean human face, which requires a short input video as the guidance. With the proposed two modules, experiments show that our method generates promising dancing videos with fewer consistency errors and best matches with human subjective evaluations.

Our main contributions are summarized as follows: 1) We propose a novel framework for synthesizing dancing videos from music sequences; 2) We propose a cross-modal alignment module to explore the relationship between music and dance clips; and 3) We propose a new method to align different pose fragments and align the rhythm of music and dance sequences.

The remainder of this paper is organized as follows: section \ref{Sec:relate} reviews related work and \autoref{sec:dataset} describes the proposed dataset. In \autoref{sec:approach} we describe our methods. We further conduct qualitative and quantitative experiments in \autoref{sec:experiment}. Finally, we conclude and discuss this paper in \autoref{sec:conclusion}.

\section{Related Work}\label{Sec:relate}

\textbf{\textbf{Pose estimation.}} Recent researches~\cite{toshev2014deeppose,wei2016convolutional,cao2017realtime,newell2017associative} tend to solve the pose estimation problem using deep learning techniques.
Toshev \emph{et al.} \cite{toshev2014deeppose} first applied the deep learning method to human pose estimation, namely deep pose, which achieves reliable performance on complex scenarios. Carreira \emph{et al.} \cite{carreira2016human} proposed a self-correcting model (iterative error feedback) to solve the problem that feed-forward neural networks cannot model dependencies in the output space efficiently. In \cite{tompson2014joint,yang2016end,ge20193d,cai2019exploiting}, researchers applied the graph model to the neural network to enhance the prediction ability of the system.
In \cite{wei2016convolutional,newell2016stacked,chu2017multi}, researchers built a multi-stage CNN regression model to expand the perceptive field of the network, so as to improve the predictive power of the model. Cao \emph{et al.} \cite{cao2017realtime} used the vector field to model the different limb structures of the human body, which solved the problem of wrong connection caused by the use of intermediate points of limbs.
A pose grammar \cite{fang2018learning} was proposed to solve the problem of 3D human pose estimation. Ge~\etal~\cite{ge2018hand} adopted the point-cloud data and proposed to regress a low dimensional representation of the 3D hand. Moreover, Cai~\etal~\cite{cai20203d} further proposed a depth-free testing model for 3D hand pose estimation.


\textbf{\textbf{Action prediction.}} Due to the variability and randomness of movements, action prediction is a hard challenge. A large body of work \cite{fragkiadaki2015recurrent,martinez2017human,habibie2017recurrent,holden2016deep,li2017auto,walker2017pose,chao2017forecasting,liu2016spatio} used LSTM to predict the next action based on a short video. Suwajanakorn \emph{et al.} \cite{suwajanakorn2017synthesizing} built a hierarchical system to predict lips movements and synthesize Obama's speech videos. For the prediction of dancing actions, earlier works~\cite{lee2013music,fan2011example,ofli2011learn2dance} proposed to learn the dancing motion based on hand-crafted features. For example, Fan~\etal~\cite{fan2011example} proposed to synthesize the motion of virtual characters based on the music style and beats rate.
Yu~\etal~\cite{yu2020structure} leveraged the self-attention mechanism to adaptively sparsify a complete action graph in the temporal space. Wang~\etal~\cite{wang2020learning} proposed to model smooth and diverse transitions on a latent space of action sequences with much lower dimensionality.
Mao~\etal~\cite{mao2019learning} and Cai~\etal~\cite{cai2020learning} converted the motion into the frequency domain by using the discrete cosine transform.
Tang \emph{et al.} \cite{tang2018dance} proposed an LSTM-based autoencoder to mapping dancing skeletons based on the musical beat. Lee~\etal~\cite{lee2019dancing} proposed a decomposition-to-composition network to align the dancing skeletons with musical beats.  We will further elaborate on the differences and relations to these works in~\secref{sec:dis}.


\textbf{\textbf{Action recognition.}} Motivated the highly accurate pose estimation techniques, it has become a trend to use skeletons and joints to classify human action. Various methods of action recognition based on the skeleton were proposed during the last decade. Driven by the success of deep learning, the skeleton modeling method based on deep learning for action recognition has emerged. Some works \cite{shahroudy2016ntu,zhu2016co,liu2016spatio} built an action recognition model based on RNN. Li \emph{et al.} \cite{li2017skeleton} and Ke \emph{et al.} \cite{ke2017new} built a model based on temporal CNNs in an end-to-end manner.  Weng~\etal~\cite{weng2017spatio} proposed a Spatial-Temporal Naive-Bayes Nearest-Neighbor for skeleton-based action recognition. Besides, Yan \emph{et al.} \cite{yan2018spatial} first applied GCN to the task of action recognition based on the skeleton, which achieves promising results in action recognition.

\textbf{\textbf{Image synthesis.}} With regard to person image synthesis, recent studies \cite{balakrishnan2018synthesizing,de2018semi,de2019conditional,joo2018generating,lassner2017generative,ma2017pose,ma2018disentangled} usually generated a character image based on a new pose. These works have achieved remarkable results in generating the details of the image. Ma \emph{et al.} proposed a new framework \cite{ma2017pose} and loss function \cite{ma2018disentangled}, but the framework did not involve audio. In \cite{villegas2017learning}, the approach has shown that pose can be used as an effective supervisory signal for prediction and video generation. Chan \emph{et al.} \cite{chan2019everybody} built an interesting model to generate a target person dancing video with the same actions as the source person video.
In this paper, we focus on the audio-inspired dancing video generation while adopting the pose skeletons as intermediaries for generation guidance.


\textbf{\textbf{Multi-modal studies.}} Many methods have achieved remarkable results in multi-modal studies. For example, Mori \emph{et al.} \cite{mori1999image} proposed a model to generate text from the image and describe the contents of the image. In \cite{tan2019text2scene,li2019object}, and \cite{yin2019semantics}, all of them extracted text features to generate a single image. Moreover, the StoryGAN \cite{li2019storygan} was proposed to generate images from text to form a sequence of images that matches the content of the text. Liao \emph{et al.} \cite{liao2015audeosynth} explored the video manipulation driven by music. In addition, Chen \emph{et al.} \cite{chen2000joint} showed face recognition would achieve better results by adding audio input. Wang \emph{et al.} \cite{wang2005inferring} also built a complex system to prove that audio input is beneficial for body pose estimation.
The system is motivated by MatchNet \cite{han2015matchnet}, which was aimed to extract the features of different objects to calculate their similarity.



\begin{figure}[t]
\begin{center}
   \includegraphics[width=1.0\linewidth]{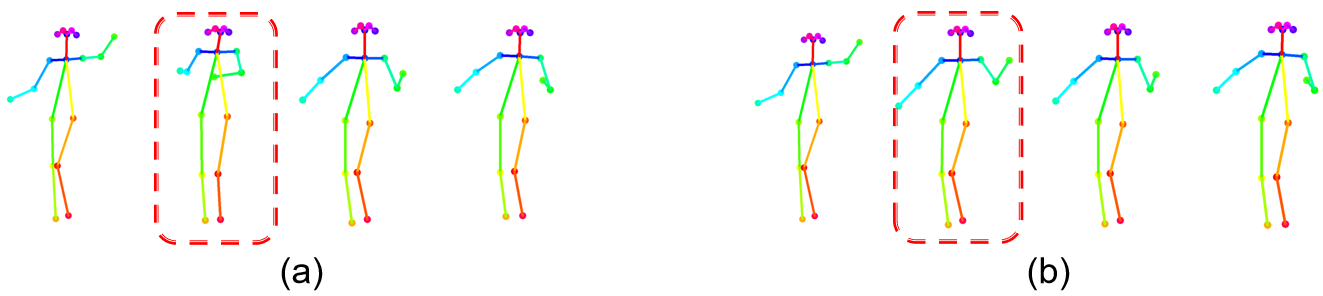}
\end{center}
   \caption{The detection result after smoothing: (a) Jitter (in red dotted box) in the process of detecting body keypoints. (b) Results after smoothing.}
\label{fig:smooth}
\end{figure}

\begin{figure}[t]
\begin{center}
   \includegraphics[width=0.8\linewidth]{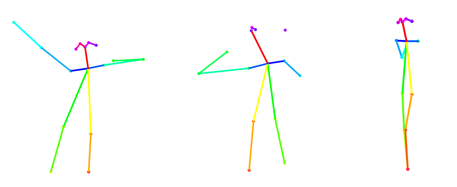}
\end{center}
   \caption{Exemplar failures of keypoint detection, which can be automatically removed before constructing training data.}
\label{fig:detecterror}
\end{figure}

\section{Dataset}\label{sec:dataset}

\subsection{Dataset Collection}\label{sec:datasetcollect}
To synthesize the dancing imaginations from the corresponding videos, we first collect the dancing videos in the real scenarios, in which the moving dancers and audios can be obtained simultaneously. As the intermediate of the dancing descriptors, we thus collect the pose sequences in an automatic manner and meanwhile constructing the database of pose fragments.
In addition, to simplify the problem formulation of generation, we tend to collect videos with only one dancer, thus the generation task of multiple people can be acquired by repeating this process.

\textbf{Audio preprocessing.}
The original audios from the real videos lie in the frequency domain, which is hard for extraction and subsequent processing. To obtain the aligned audio features with video frames, we adopt the MFCC algorithm~\cite{zheng2001comparison} to transform the frequency signals to low-dimension vectors. Specifically, we compute the features\footnote{https://github.com/jameslyons/python\underline{~}speech\underline{~}features} and choose the length of the windows as $1,000/video_{fps}$ with $fps=24$,~\ie, $41.66$ms. The final audio feature dimension is $13$ which can be easily embedded in our system.

\textbf{Pose extraction.}
As mentioned above, we need to extract the body keypoints from dancing videos in our system. Instead of the manual labeling of each video frame, we automatically obtain human keypoints by performing the OpenPose \cite{cao2018openpose} algorithm. We find that there is an accidental detection error such as the local jitter (see, \figref{fig:smooth} (a)) in the continuous frames. To smooth the pose sequences and filter the out-liners, we conduct a simple linear interpolation among different pose frames. The smoothed results are shown in \figref{fig:smooth} (b). In addition, we automatically delete the incomplete detection (see, \figref{fig:detecterror}) following two rules: 1) it and the successive frames that follow it are too far from the previous frame, or 2) hand and foot nodes are not detected.
After filtering the abnormal data, we segment the video clips as fragments with non-overlapping sliding windows,~\ie, 1s to 4s.
We divide the extracted pose sequence into different fragments in seconds and then construct the fragments into the sequence database $\mathbbm{D}$.

\begin{figure}[t]
\begin{center}
   \includegraphics[width=0.9\linewidth]{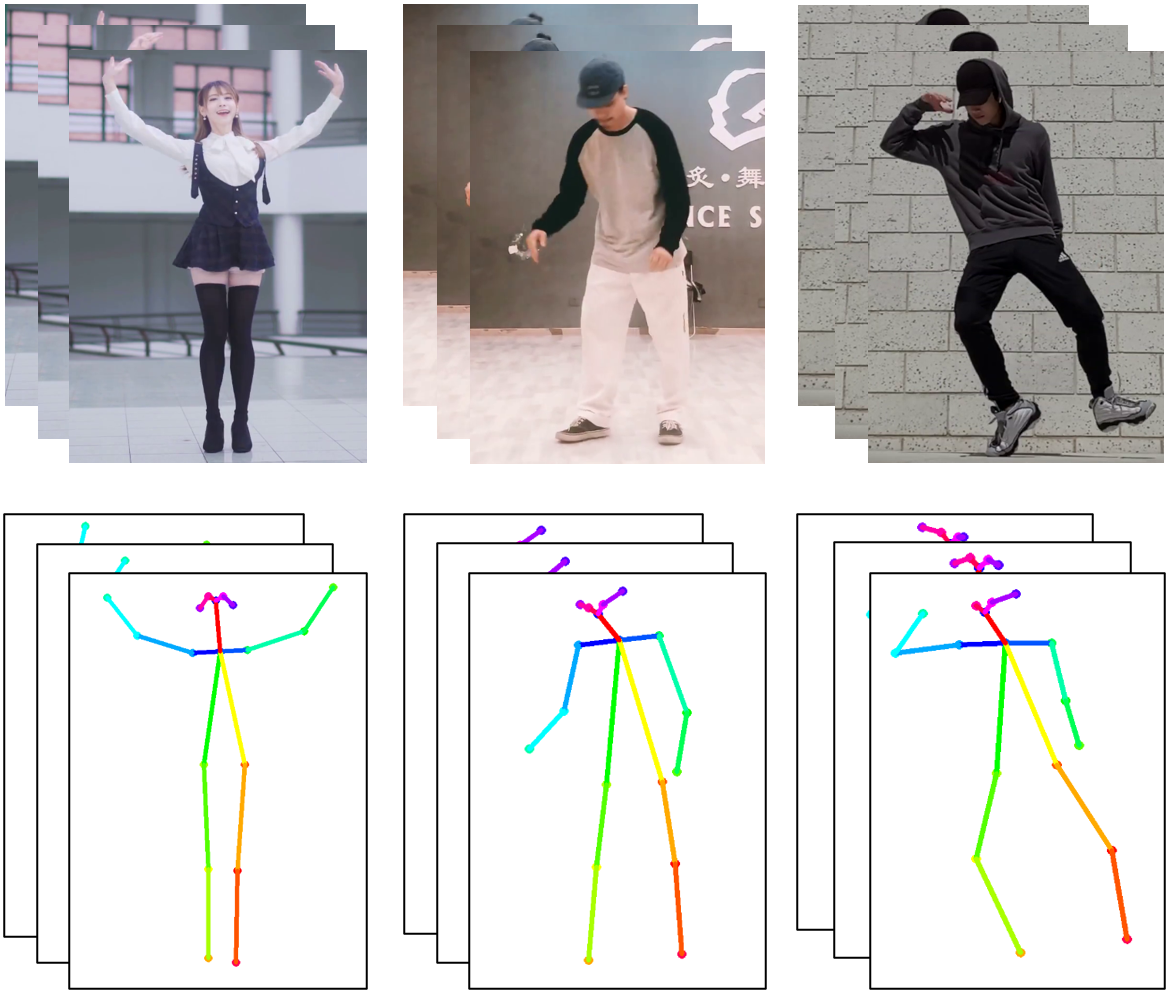}
\end{center}
   \caption{Examples of our dataset: video frames in the dancing video and their corresponding pose keypoints.}
\label{fig:dataset}
\end{figure}

\begin{figure}[t]
\begin{center}
   \includegraphics[width=1\linewidth]{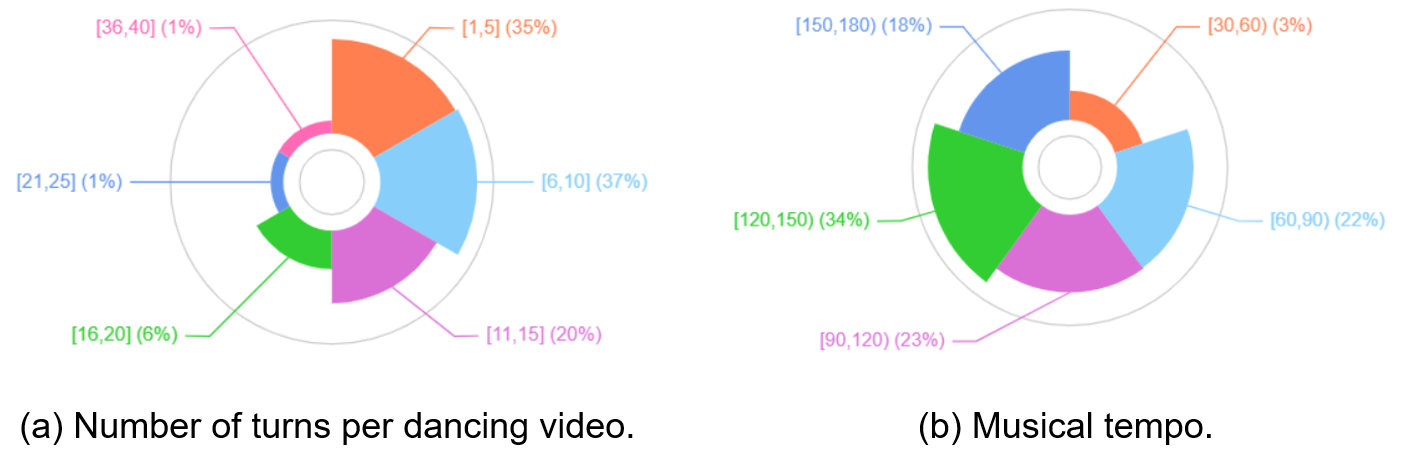}
\end{center}
   \caption{Dataset statistics. We extract the tempo of each music and the number of turns in each video from the collected dance dataset. Then we present the distribution of the number of turns and musical tempo.}
\label{fig:turn}
\end{figure}

\subsection{Dataset Statistics}

\textbf{Data scale.} We collect 122 dancing videos of females and 32 videos of males, including ACGN, robot, and hip-hop dance. Our collected videos range from 3 min to 5 min. A total of 9.0 hours of dancing videos is collected. We process all videos to $24fps$, extract frames into the dance image set and resize them to the same size. The total number of frames is $706,463$.

\textbf{High quality.} To construct a high-quality dataset, the dancing videos we collected are high-resolution ($1920\times1080$) and have a stable fixed camera and bright lighting. Specifically, we select videos with high-quality music sound, noiseless interference, and solo performance. \figref{fig:dataset} shows example frames from videos and their corresponding keypoints.

\textbf{Abundant diversity.} As shown in \figref{fig:turn}, we count the number of turns and the musical tempo in each dancing video. The tempo of the music and the number of turns vary significantly in each video. Moreover, the number of turns in all videos is greater than 1, and most of the number is greater than 5 (see, \figref{fig:turn} (a)).


\begin{figure*}[t]
\begin{center}
   \includegraphics[width=1.0\linewidth]{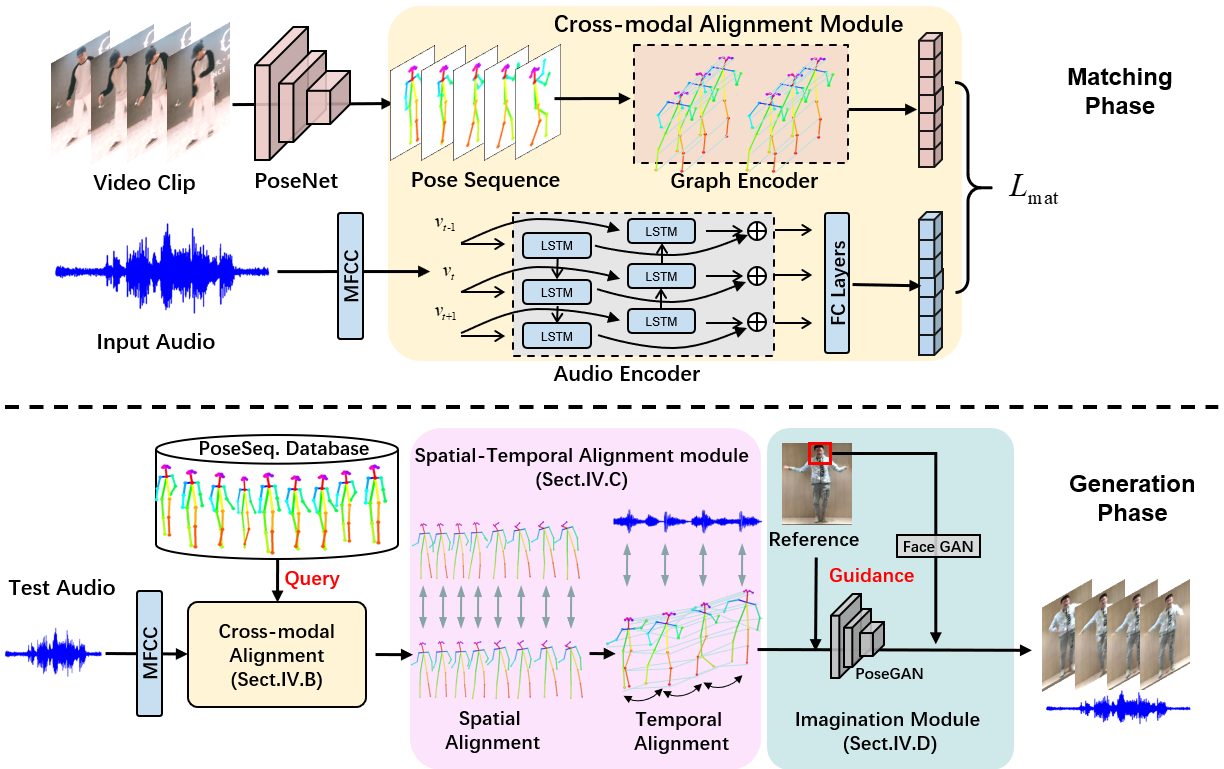}
\end{center}
   \caption{The pipeline of the proposed framework.  Matching Phase: given input video and audios, the matching phase learns the relationship between these two different modalities. Generation Phase: the spatial alignment is used to re-predict dance movements between discontinuous pose fragments. The temporal algorithm aligns the beats of the music and pose sequences. The imagination module synthesizes the final dancing videos from the processed pose sequences.}
\label{fig:method}
\end{figure*}

\section{The Approach}\label{sec:approach}

\subsection{Overview}

Given an input audio clip $\mc{A}$ and source video frames $\mc{V}_s$, the target of our synthesis framework is learning to generate target video frames $\mc{V}_t$, while adopting the corresponding pose features $\mc{P}$ as an intermediary. Let $\{\textbf{x}^s_i=(\textbf{v}^s_i,\textbf{a}^s_i,\textbf{p}^s_i) | \textbf{x} \in \mathbb{S} \}_{i=1}^{N}$, which is a triplet of video, audio, and pose frames from the input source domain $\mathbb{S}$. And the triplet $\textbf{x}^t \in \mathbb{T}$ of target domain is defined accordingly. Divided by the source and target domain, our framework can be constructed in two phases, which are shown in~\figref{fig:method}.

In the first matching phase, we take the generated $\textbf{x}^s$ as input and thus regularize the cross-modal alignment $\Phi_c(\cdot)$ to learn the correspondence between each $\textbf{p}^s_i$ and $\textbf{a}^s_i$. Instead of direct learning the relation of specific identities in the limited source video, we expound this audio-to-video correspondence into a general condition by measuring the distance between audios and pose sequences. While pose features are usually regarded as the guidance for human video generation. This relationship learning process has the form:
\begin{equation}
\theta_c=\arg\min_{\theta} \Phi_c(\textbf{a}^s_i,\textbf{p}^s_j | \theta)\mathbbm{1}(i=j)-\Phi_c(\textbf{a}^s_i,\textbf{p}^s_j | \theta)\mathbbm{1}(i \neq j),
\end{equation}
where $\mathbbm{1}(\cdot)$ is the characteristic function and $\theta$ denotes the network parameters of cross-modal alignment module $\Phi_c$.
This objective regularizes audios and videos from the same frame with closer distances and vice versa.

With the learned relationship of videos and audios, the generation phase only takes audio $\textbf{a}^t$ from the target domain as input and retrieves the nearest pose sequence $\hat{\textbf{p}}$ from the constructed sequence database $\mathbbm{D}$ (described in~\secref{sec:datasetcollect}) from the source domain. It can be formally represented as:
\begin{equation}\label{eq:optimial}
\hat{\textbf{p}}=\arg\min_{\textbf{p} \in \mathbbm{D}} \Phi_c(\textbf{a}^s, \textbf{p} | \theta_c).
\end{equation}
The retrieved pose sequence $\hat{\textbf{p}}$ is usually fragmented, causing the generated dancing movements unnatural. Hence we propose a spatial-temporal alignment module $\Phi_{st}$, which refines the sequence from two different aspects. In the spatial domain, we detect the sudden changes among adjacent frames and smooth the results by an interpolation operation. In the temporal domain, we align the musical beats with substantial body movements, which makes the dancers dance rhythmically.
After the alignment module $\Phi_{st}$, the pose sequences are further encoded along with the guidance video clip $\textbf{v}_g \in \mathbbm{T}$ from the target domain to generate imagined dancing movements, which is supervised with an adversarial training process.
\begin{equation}
\textbf{v}^t=\mc{G}(\Phi_{st}(\hat{\textbf{p}},\textbf{a}^t),\textbf{v}_g),
\end{equation}
where $\mc{G}$ denotes the generation network in~\secref{sec:methodimagine}.
In the following subsections, we will elaborate the cross-modal alignment $\Phi_c$ in \secref{sec:methodcross}, spatial-temporal alignment module $\Phi_{st}$ in \secref{sec:methodspatial}, and the imagination module in \secref{sec:methodimagine}.

\subsection{Cross-modal Alignment}\label{sec:methodcross}

To measure the distance between audio $\textbf{a}^s$ and video frames $\textbf{v}^s$, we first utilize the extracted pose sequences in~\secref{sec:dataset} and corresponded audio clips and fed these encoded features to the cross-modal alignment module (view in yellow in~\figref{fig:method}).
However, these features are raw data that are simply transformed from the image or frequency domains, making the distance hard to measure.
Hence we adopt the audio encoder and the pose encoder to embed different modalities into the same latent space and finally use a metric learning optimization to learn this correlation.

\textbf{Audio Encoder.} We first transform the audio with MFCC and process in~\secref{sec:dataset} to a $13$d vector per frame.
Inspired by previous methods, which achieves satisfactory results to predict finger movements \cite{shlizerman2018audio} and lip synthesizes \cite{suwajanakorn2017synthesizing}, we adopt a bidirectional-LSTM (Bi-LSTM) and add a fully connected layer $\mathtt{fc}$ to extract audio features. With the bidirectional information flow, our encoder can extract both forward and backward information to aggregate richer features from the audio. The encoding operation of the $i$th frame is:

\begin{equation}
\stackrel{\rightarrow}{h_i} = \stackrel{\rightarrow}{\mathtt{LSTM}}(\textbf{a}^s_i, \stackrel{\rightarrow}{h}_{i-1}, v_i),
\end{equation}
\begin{equation}
\stackrel{\leftarrow}{h_i} = \stackrel{\leftarrow}{\mathtt{LSTM}}(\textbf{a}^s_i, \stackrel{\leftarrow}{h}_{i+1}, v_i),
\end{equation}
\begin{equation}
\hat{\textbf{a}}^s_i = \mathtt{fc}(\stackrel{\rightarrow}{h_i},\stackrel{\leftarrow}{h_i}),
\end{equation}

where $v_i$ is the hidden layers and $\hat{\textbf{a}}^s$ is the encoded output. The parameters of Bi-LSTM that we used are hidden state of $100$, dropout of $0.1$, learning rate of $1e-3$. The final audio feature is $16$ dimensions.

\textbf{Pose Encoder.} To extract the context information of long-term frames, the generated pose sequences $\textbf{p}^s$ are then fed into a graph encoder. We adopt the ST-GCN \cite{yan2018spatial} to exploit the relationship of different pose skeletons, which thus forms a holistic understanding of each pose fragment.
The graph encoder extracts the temporal information $g_{te}$ on 2d convolution kernels, and extracts their spatial information $g_{sp}$ on graph convolution kernels:
\begin{equation}
g_{te} = \textbf{W}_c \textbf{p} + b_1,
\end{equation}
\begin{equation}
g_{sp} = \Lambda^{-\frac{1}{2}}(\textbf{A} + \textbf{I})\Lambda^{-\frac{1}{2}}g_{te} \textbf{W}_g,
\end{equation}
\begin{equation}
\hat{\textbf{p}}_t = \mathtt{fc}(g_{sp}) + b_2,
\end{equation}
where $\textbf{W}_c$ and $\textbf{W}_g$ represent 2d convolution kernel and the graph convolution kernel respectively. $\textbf{A}$ is the adjacent matrix for the pose and $b_{\{1,2\}}$ are the learnable bias. Here $\Lambda^{ii}=\Sigma_j(\textbf{A}^{ij}+\textbf{I}^{ij})$. The embedded features of pose $\textbf{p}^s$ and audio $\textbf{a}^s$ are encoded with the same size of $16$d.

\begin{figure}[t]
\begin{center}
   \includegraphics[width=1.0\linewidth]{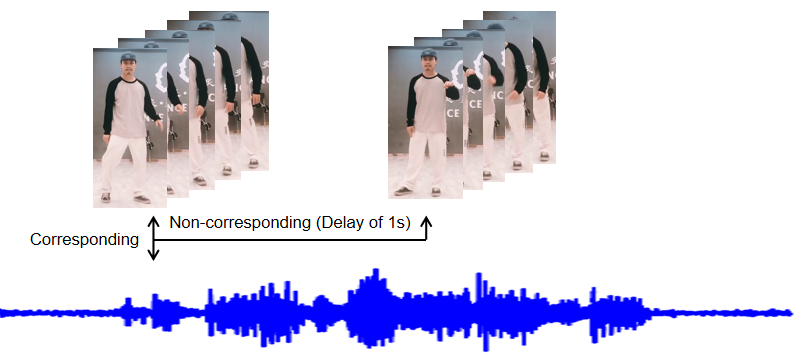}
\end{center}
   \caption{Example of training data. Positive samples: corresponding audio with pose fragments. Negative samples: non-corresponding audio with delayed pose fragments.}
\label{fig:delaycase}
\end{figure}

After obtaining music features and corresponding dance features, we calculate the similarity by Euclidean distance. We propose a correlation matching loss $\mc{L}_{mat}$ to regularize the similarity of features. The encoded pose feature $\hat{\textbf{p}}$ and audio feature $\hat{\textbf{a}}$ are attached with closer distance if they are from the same sequence, and vice versa. The overall loss function can be formulated as:
\begin{equation}\label{eq:loss}
\mc{L}_{mat} = ||\hat{\textbf{p}}_i-\hat{\textbf{a}}_j||_2^2 \mathbbm{1}(i = j) +||\hat{\textbf{p}}_i-\hat{\textbf{a}}_j- \epsilon||_2^2\mathbbm{1}(i \neq j).
\end{equation}
If $\hat{\textbf{p}}$ and $\hat{\textbf{a}}$ are relevant features, the loss function calculates the distance of two features and feeds it back to the network. On the irrelevant condition, we add a parameter $\epsilon$ to enlarge the distance of two features.
However, only learning from the corresponding cases can easily lead to over-fitting. To solve this problem, we add a non-corresponding case to training data, where the pose fragments are delayed by a few seconds than MFCC features, which is illustrated in~\figref{fig:delaycase}.

\subsection{Spatial-Temporal Alignment}\label{sec:methodspatial}
With the learned MatchNet $\Phi_c (\theta_c)$, each test audio could retrieve a sequence of pose fragments with the optimization in Eqn.~\eqref{eq:optimial}. Although the retrieved pose fragments generate reliable movements, there exists a severe misalignment between different pose fragments, considering the fragments are discretely sampled in the construction of database $\mathbbm{D}$. Toward this end, it is desirable to further rearrange the sequence with a spatial-temporal alignment module to make the dancing sequence vivid.

\textbf{Spatial alignment.} In the generation phase of~\figref{fig:method}, the audio guidance $\textbf{a}^t$ from the target domain is encoded in the same way to get an audio vector. Instead of predicting the pose of each frame, we retrieve a most similar pose sequence $\textbf{p}^t$ with the built database $\mathbbm{D}$. However, it occurs abnormal movements among multiple pose fragments. Thus to solve this problem, we first adopt a spatial alignment to align the features being reasonable.
The first step of spatial alignment is to detect the discontinuous frame in the retrieved pose sequences, whose skeletons show jitters with the previous frame. We define one frame as the discontinuous frame whose keypoint movements are larger than 10 pixels. Our basic idea is to refine these out-liners with a smooth movement function, which can be estimated with the whole dancing sequences.

In practice, the music melodies are composed of ``beats", which contain periodic signals but with dynamic changes in each period.
Owing to the seq-to-seq design, here we introduce the temporal series decomposition for aligning and smoothing the possible non-continues frames between adjacent sequences.
Taking this abnormal frame $p_k$ as the central frame, we thus select the frames in nearby window $ i \in \lceil k-\omega_a / 2, k+\omega_a / 2 \rfloor$ to form a Time Series Decomposition (TSD) algorithm.

With the defined sliding window, one notable issue is retaining the position of the start point and endpoint unchanged, which guarantees consistency after the overall refinement.
Therefore we resort to a linear fitting operation $\mathcal{R} = \mathcal{F}(\br{p})$ from the startpoint and endpoint of this sequence and $\mathcal{F}(\br{p}_i)$ represents the $i$th value on this line. Hence we take the misalignment $\br{d}_i=\br{p}_i-\mathcal{F}(\br{p}_i)$ for optimization. Once abnormal frames are detected, we conduct this smoothing operation inside each window $ i \in \lceil k-\omega_a / 2, k+\omega_a / 2 \rfloor$.

Following the basic idea of TSD, we simply decompose each keypoint misalignment $\br{d}$ with the trend term $M$, the periodic term $S$ and a random term $\gamma$. Taking one keypoint misalignment $\br{d}_i$ of the $i$th frame as an example, it has the form:
\begin{equation}
\textbf{d}_i = S_i + M_i + \gamma_i.
\end{equation}
 We then perform differential operations $\varphi(\cdot)$ on $\textbf{d}$ to eliminate the periodic movement and obtain the tend function. The differential displacement $\hat{\br{d}}_i$ without periodicity can be formulated as:
\begin{equation}
\begin{split}
\hat{\br{d}}_i =  M_i(i | \alpha) = \alpha_0 + \sum_{m=1}^3\alpha_m i^m,
\end{split}
\end{equation}
where in each period $T$, the trend term $M$ can be fitted with the learnable weight $\alpha$ of the cubic polynomial function.

The periodic term $S$ is to find the minimum period $T$ for $S_t=S_{i+T}$ and $T$ is determined by volatility threshold $th=5$ pixels.
\begin{equation}
 S_t=\mathbb{E}(\br{d}_i- \hat{\br{d}}_i),
\end{equation}
where $\mathbb{E}(\cdot)$ denotes the average expectation.
The stationary random term $\gamma \sim \mc{N}(0,\sigma^2)$ is normally distributed.
With this learnt decomposition, our target is using the previous $\omega_b (\omega_b>\omega_a)$ frames as reference frames and conduct this alignment per $\omega_a$ frames. We thus predict the TSD parameters of the next $\omega_a$-length window $(M^*,S^*,\gamma^*)$ and update the final coordinate point $\widetilde{\textbf{p}}$, which is illustrated in~\algref{algorithm1}.

\begin{algorithm}[h]

  \caption{Spatial Alignment Algorithm} \label{algorithm1}
  \SetKwInOut{Input}{Input}\SetKwInOut{Output}{Output}
 \Input  {Target pose sequence $\textbf{p}^t$: a sequence of body keypoint coordinates. \\ TSD parameters: $\{(M_i,S_i,\gamma_i), i=0 \ldots \omega_b \}$.}
 \Output {Updated coordinate point: $\widetilde{\textbf{p}}$.}
 \Begin{
        {\bf Initialization: $\varphi(\cdot)$ is the difference equation and $\xi(\cdot)$ calculates the length of the data in a period.}\\
\Repeat{i $\geq$ $\omega_b$}{

Make the linear fitting between the start and end points:\\
$\mathcal{R} = \mathcal{F}(\textbf{p})$\\
Calculate the relative displacement of points in this window:\\
$\textbf{d} \longleftarrow \textbf{p} - \mathcal{F}(\textbf{p})$\\
Establish the spatial-temporal equation based on periodicity and trend:\\
$\textbf{d}_i = M_i + S_i + \gamma_i$\\
Estimate and extract $M_i$ and $S_i$:\\
$\hat{\br{d}}_i = \varphi(\textbf{d}_i)$\\
$M_i = \arg\min\limits_{\alpha}(\hat{\br{d}}_i-\sum_{m=0}^{3}\alpha_m i^m)^2$\\
$S_i = \mathbb{E}(\br{d}_i- \hat{\br{d}}_i)$\\
$i \longleftarrow i+\omega_a$
}
Predict $M^{*}$ and $S^{*}$ of the abnormal window:\\
$M^{*} \longleftarrow \mathbb{E}(M_i)$\\
$S^{*} = \arg\max\limits_{S_i}(\xi(S_i))$\\
Make the linear fitting between the start and end points:\\
$\mathcal{R} = \mathcal{F}(\textbf{p}^{*})$\\
$\textbf{d}^{*} = M^{*} + S^{*} + \gamma^{*}$\\
$\widetilde{\textbf{p}} \longleftarrow \textbf{d}^{*} + \mathcal{F}(\textbf{p}^{*})$

}
\end{algorithm}

\textbf{Temporal alignment.} Although the dancing sequences are well-smoothed by the spatial alignment, it seems unnatural when the musical beats (abrupt changes in audios) are not aligned with the large motions,~\eg, raising hands. Thus we adopt a temporal alignment to make the musical beats and motions consistent.

To make the sequences align with the music beats, we first extract the beat points of the target audio $\br{a}^t$ with the \textit{librosa} library~\cite{mcfee2015librosa}.

In the pose sequence, these pose beats are defined by:
\begin{equation}\label{eq:temporal}
\mu = \arg\max\limits_{j}(|\widetilde{\textbf{p}}_j^t-\widetilde{\textbf{p}}_{j-1}^{t}|), j\in \lceil i, i+\omega_c \rfloor.
\end{equation}

 We select $\omega_c$ adjacent frames as a sliding window to search the local maximum to align.
 With the detected audio beats $\br{a}_{\mu}$, the $\omega_c$ frames can be re-organized as $[i, \mu]$ and $(\mu, i+\omega_c]$.
 For the preceding and the following frames of the current beat point, we adopt cubic fitting interpolation to get the aligned pose $\hat{\textbf{p}}^t$. This align function then has the form:
\begin{equation}
\alpha^*=\arg\min\limits_{\alpha}\sum_{x=i}^{\mu}(\sum_{m=0}^{3}\alpha_m x^{m}-\widetilde{\textbf{p}}_x^t)^2,
\end{equation}
\begin{equation}
\hat{\textbf{p}}^t=\alpha_0^*+\sum_{m=1}^{3}\alpha_m^*x^m.
\end{equation}
The optimization of $(\mu, i+\omega_c]$ is conducted in the same manner. After this temporal alignment, the musical beats and abrupt pose movements can be well aligned, which can serve as guidance for video imagination.


\subsection{Imagination}\label{sec:methodimagine}
Taken the optimized $\hat{\textbf{p}}^t$ and video guidance $\textbf{v}_g$ from target domain $\mathbbm{T}$, the final imagination is to generate reliable video frames.
To achieve this goal, we adopt the pix2pixHD~\cite{wang2018high} as the generator for poseGAN. Video guidance $\textbf{v}_g$ is a 2-minute length video with the target character acting simply. In the stage of training, the goal of training generator is to generate a more realistic image and fools discriminator to regard as real. On the contrary, the purpose of training discriminator is to identify whether the image is from the real images or the generator synthesizing.

However, there still exist some abnormal regions in the generated images, especially in the face region. To enhance the representation of face regions, we adopt the FaceGAN \cite{chan2019everybody} to enhance the facial region of the generated image. The purpose of the generator in the FaceGAN is to synthesis the detailed supplement of the face area between the synthesized images and the real images. We define the face region as the area near the nose position (i.e. 50 $\times$ 50 patches centered around the nose node) in the detected body pose, which is illustrated in \figref{fig:facegan}. The generator of the FaceGAN outputs a residual gradient to help to synthesize the final face region. Additionally, in the process of training FaceGAN, the purpose of the discriminator is to distinguish the image synthesized by the generator or from the real images.

\subsection{Discussions and Relations}\label{sec:dis}
To make a theoretical analysis with state-of-the-art models~\cite{tang2018dance,lee2019dancing}, here we elaborate on three major differences as follows.

 \textbf{Relationship learning.} Both Lee~\etal's method~\cite{lee2019dancing} and Tang~\etal's method~\cite{tang2018dance} are established based on the direct deduction from input music to dancing skeletons. However, direct mapping one modality to the other would usually lead to unreliable results. To this end, our work proposes a new perspective to explore the relationship between input audio sequence to a dancing sequence, namely seq-to-seq mapping. This guarantees the smoothness and rationality of the generated dancing videos.

  \textbf{Audio understanding.} Tang~\etal~\cite{tang2018dance} proposed to learn the skeleton movement with only extracting the ``beat'' frames to retain the motion alignments, which neglects the contextual information of input audio. To make a further improvement, Lee~\etal~\cite{lee2019dancing} adopt a music style extractor to obtain the whole artistic style of the audio content,~\eg, sadness or happiness. While in these researches, the contextual understanding of musical information can be easily neglected. By contrast, our proposed approach extracts the global information of the whole input audio, maintaining the audio content for the cross-modal relationship learning.
     We believe that understanding the whole audio content but not only the ``beat" is necessary for the dancing video synthesis.

   \textbf{Generation targets.} The generation target of~\cite{lee2019dancing} and~\cite{tang2018dance}  are dancing skeletons,~\ie, human body keypoints. However, the final motif of music-inspired synthesize is to generate vivid and natural dancing videos. Hence, in our method, an imagination module is proposed to synthesize the dancing videos, and the skeletons play an intermediate role in the generation framework.

\section{Experiments}\label{sec:experiment}

\subsection{Experimental setting}
To quantitatively evaluate the generated dancing results of our approach, we conduct experiments from two aspects,~\ie, the quality of pose sequence and synthesized videos\footnote{Some results can be found at \textsuperscript{}http://cvteam.net/projects/danceit/results.mp4 }. To evaluate the generation of pose sequences, we randomly select $80\%$ of the data for training and the rest $20\%$ for testing. In the processing of dataset assembling (\secref{sec:datasetcollect}), we remove about $10\%$ of the frames in the videos due to the extracted pose data that cannot be tackled.
To synthesize dance videos in the imagination module, we collect 8 videos $\textbf{v}_g$ with a length of 120 seconds from different participants. We sample these generation videos in different natural scenarios, which have different backgrounds, clothing, and illuminations.
\subsection{Implementation details}

Attributed to the light-weight implementation, our whole framework can be trained on a single consumer-level NVIDIA RTX 2070 GPU, $12$ cores CPUs, and 16GB memory. It takes $35.1$ hours for detecting pose keypoints and $0.5$ hours for extracting the MFCC features of the music. The running time for 500 epochs of training of cross-modal system takes $2.9$ hours (20.9s per epoch). After getting videos from the participants, we cost $12.0$ hours to train the poseGAN to generate the target video image from the pose. In addition, it takes $0.7$ hours to train the FaceGAN, which is used to strengthen facial areas.
Since the shortest pose fragment is $1$s and the number of frames is $24$, we thus adopt $\Phi_{st}$ with $\omega_a$ of $8$ and $\omega_b$ of $24$ to conduct the experiments.

\begin{table}[t]

\caption{
Quantitative comparisons of our proposed method and Lee~\etal~\cite{lee2019dancing}. $\mc{S}_{BA}$ denotes the beat alignment rate. The best performances are in bold. $\uparrow$: the higher the better. $\downarrow$: the lower the better.}\label{tab:comparision}
\setlength{\tabcolsep}{1.8mm}
\begin{center}
\begin{tabular}{c|cccccc}
\toprule
Methods & $\mc{S}_{BA}$$\uparrow$&MDD-H$\downarrow$&MDD-F$\downarrow$&SDD-H$\downarrow$&SDD-F$\downarrow$\\
\midrule
Lee~\etal~\cite{lee2019dancing}&60.05$\%$ & 8.17& 7.05 & 6.68 & 6.09\\
Ours (2s)& 76.25$\%$ & 6.23 & 5.67 & 6.81 & 5.99\\
Ours (3s)& 67.93$\%$ & 5.84 & 5.22 & 6.98 & 6.07\\
Ours (4s)& \textbf{78.80$\%$}&\textbf{5.63}&\textbf{5.17}&\textbf{6.67}&\textbf{5.97}\\
\bottomrule
\end{tabular}
\end{center}

\end{table}

\begin{table}[t]

\caption{
Hand and foot moving distribution comparisons with state-of-the-art. (x1,x2) represents the pixel distance of the moving range between adjacent frames. The values are shown with the percentage of the overall distribution.
}\label{tab:hand}
\begin{center}
\begin{tabular}{c|ccccc}
\toprule
Methods &(0,20)&(20,40)&(40,60)&(60,80)&(80,400)\\
\midrule
\multicolumn{4}{c}{Hand Moving Distribution}  & &     \\
\cline{1-6}
Lee~\etal~\cite{lee2019dancing} &91.0973 & 8.4076 & 0.4412 & 0.0322 & 0.0110 \\
Ours& 95.6649 & 4.2475& 0.0867 & 0.0008 & 0 \\
\midrule
\multicolumn{4}{c}{Foot Moving Distribution}  & &     \\
\cline{1-6}
Lee~\etal~\cite{lee2019dancing} &98.7880 & 1.0918 & 0.1022 & 0.0046 & 0.0026 \\
Ours& 98.7214 & 1.2640 & 0.0134 & 0.0012 & 0  \\
\bottomrule
\end{tabular}
\end{center}

\end{table}

\begin{figure}[t]
\begin{center}
   \includegraphics[width=1.0\linewidth]{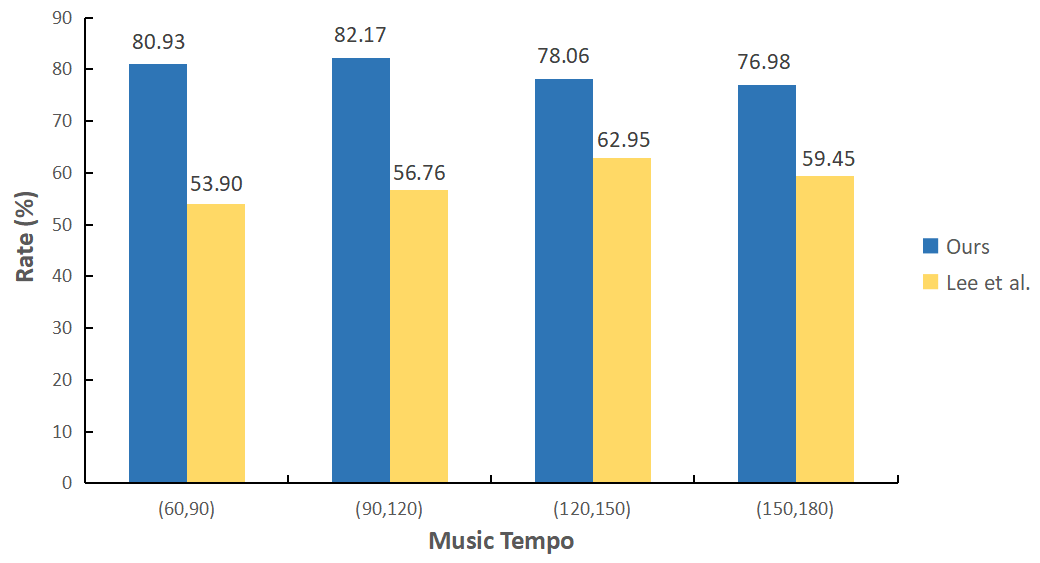}
   \caption{
   Beat alignment score $\mc{S}_{BA}$ under different musical beat environment. The tempo values are presented by beat number/per minute.}
\label{fig:beat}
\end{center}
\end{figure}

\subsection{Comparison with State-of-the-art}

\subsubsection{Comparisons of skeleton alignment}
We adopt the official model of~\cite{lee2019dancing} and test our model and~\cite{lee2019dancing} with the same input test. Here we propose three metrics to evaluate the quality of generated skeletons.

\textbf{ Beat alignment score:}
A musical beat is defined as the maximum value of body movement in each sequence unit in Eqn.~\equref{eq:temporal}. Hence our aim is to calculate the matched beat number between the input audio $\br{a}^m_t$ and body movement $\mu$:
\begin{equation}
    \mc{S}_{BA} = \frac{\sum_{m=0}^{n} \mathbbm{1}(\br{a}_m^t = \mu_{m}) }{B_n},
\end{equation}
 where $\mathbbm{1}(\cdot)$ denotes the indicator function and $B_n$ denotes the total number of the beat. The measurement $\mc{S}_{BA}$ tends to be higher if there is a synchronization of input music and generated dancing movement.

\textbf{Hand/Foot Moving Distribution Distance (MDD):}
 Audio-inspired synthesizing methods usually generate different dancing movements when encountering the same input audio. Hence, we proposed to measure the skeleton movement between adjacent frames.~\ie, human beings would present similar \textit{moving range} when hearing similar music.
 We measure the distance between hand and foot distributions between the groundtruth source $\br{p}^{s}$ and estimated pose $\hat{\br{p}}^{t}$ as follows:
 \begin{equation}\label{eq:mdd}
 MDD_{\{h, f\}} = \frac{\bs{KL}_{\{h,f\}}(\Delta \hat{\br{p}}^{t} || \Delta \br{p}^{s})+\bs{KL}_{\{h,f\}}(\Delta \br{p}^{s} || \Delta \hat{\br{p}}^{t})}{2},
 \end{equation}
 where $MDD_{\{h, f\}}$ denotes the hand and foot moving distribution distance respectively and $\bs{KL}(\cdot)$ represents the KL divergence scores. $\Delta$ denotes the difference between two adjacent frames,~\ie, the difference between $\br{p}_{i}$ and $\br{p}_{i-1}$.

\textbf{Hand/Foot Spacing Distribution Distance (SDD):} Besides the movement of hand and foot keypoints, here we propose a new measurement to describe the spacing distance of one person,~\eg, the distance between one's feet. These measurements are defined similarly using Eqn.~\eqref{eq:mdd} to describe the discrepancy of adjacent frames.

\begin{figure*}[t]
\begin{center}
   \includegraphics[width=1\linewidth]{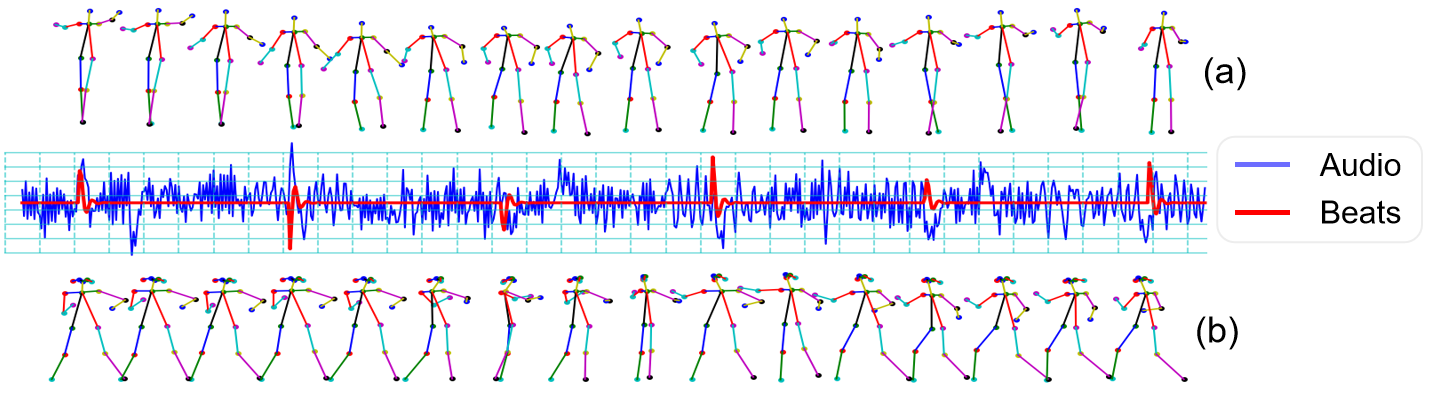}
\end{center}
   \caption{ Visualizations of beat alignment attribute. From top to bottom: generated pose sequences of Lee~\etal~\cite{lee2019dancing}, input audio guidance, our generated pose sequences.}
\label{fig:2DComparison1}
\end{figure*}

Moreover, we exhibit the beat alignment score $\mc{S}_{BA}$ in~\figref{fig:beat} under different musical beat environment,~\ie, beat number/per minute. The results indicate that under different beat rate conditions, our proposed method shows better alignment than the state-of-the-art model~\cite{lee2019dancing}, especially in music with slow rhythm (\ie, smaller than 120 beats/per minute).

\subsubsection{Comparisons of moving distribution}
One severe problem lies in the direct-mapping strategies,~\eg,~\cite{tang2018dance} and~\cite{lee2019dancing}, is the abnormal jittering. Considering the realistic dancing video, the movement of each body keypoint would present in a normal range. A large movement between adjacent frames (\ie, 1/30s) is not encouraged. Here we present the distribution of hand/foot keypoint movements in~\tabref{tab:hand}. It can be found that our model generates more smooth and progressive body movement while \cite{lee2019dancing} performs large jittering movements. In addition, for moving distance larger than 80pixs (which is usually impossible for body motion in 1/30s), \cite{lee2019dancing} shows some abnormal distributions in both hand and foot motions. Benefiting from the sequence to sequence mapping and the spatial-temporal alignment module, our approach generates notably better results.

\begin{figure}[t]
\begin{center}
   \includegraphics[width=0.6\linewidth]{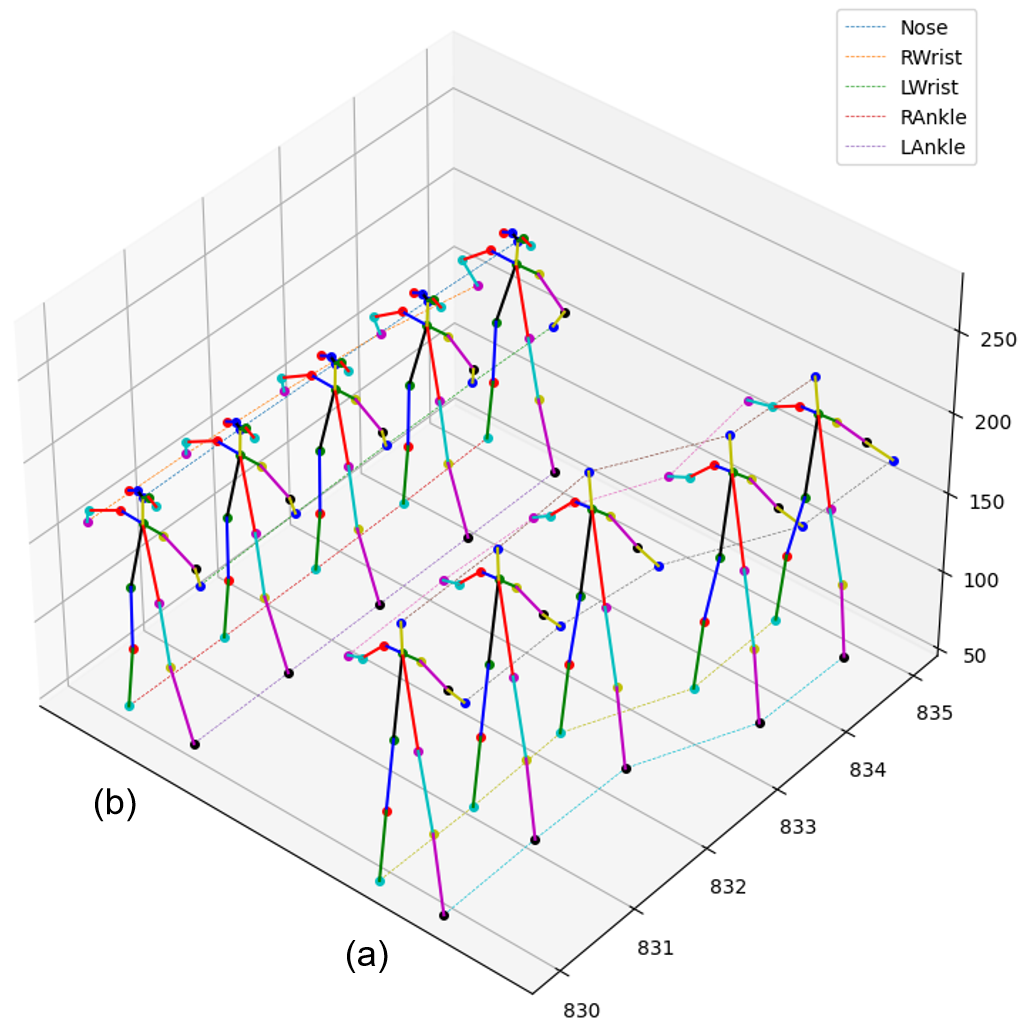}
\end{center}
   \caption{
   3D Visualizations of continuous movements. (a): skeletons generated by our proposed network. (b):skeletons generated by~\cite{lee2019dancing}.
   The three axes indicate the $x,y$ pixel localization and frame $i$.}
\label{fig:3DComparison}
\end{figure}

\subsubsection{Visualization comparisons}
To visually compare with state-of-the-art models \cite{lee2019dancing}, we exhibit four representative cases to verify the generation quality of our approach.
With the same music guidance in the bottom (musical beats are colored in red line), the generated pose skeleton and our skeleton are presented at top and bottom respectively.
We exhibit a representative case with the same input music in~\figref{fig:2DComparison1}
our generated dancing skeleton has a strong ability for attaining rich body motions that are highly aligned with the musical beats. While the Lee~\etal ~\cite{lee2019dancing} in the first row shows slightly body movements but containing more abnormal jitterings.
Moreover, it can be found that our generated pose sequences show continuous rhythmic dancing sequence while \cite{lee2019dancing} shows slight changes. Benefiting from the cross-modal alignment and rich pose skeleton representation with head keypoints, our approach generates dynamic head poses while \cite{lee2019dancing} are not endowed.
Last but not least, in~\figref{fig:3DComparison}, considering the different mapping manners (\ie, direct mapping and seq-to-seq mapping), \cite{lee2019dancing} shows abnormal movement in the $834$th frame, while our model shows continuous dancing skeleton sequences.

\begin{table}[t]
\begin{center}
\caption{
The Euclidean distance (Euclid.) and correlation accuracy (Corr.) of different hyper-parameters of cross-modal alignment. $D_f$ denotes the dimension of output features.}\label{tab:hyper}
\setlength{\tabcolsep}{1.5mm}
\begin{tabular}{l|c c c c}
\toprule
Method & Train Euclid. & Test Euclid. & Train Corr. & Test Corr.\\
\midrule
$lr=0.0005$ & 2.49 & 1.79 & 0.65 & 0.73 \\
$lr=0.001$ & 4.63 & 3.50 & 0.60 & 0.70 \\
$lr=0.01$ & 16.91 & 12.46 & 0.50 & 0.65 \\
$dropout=0.2$ & 3.79 & 2.62 & 0.60 & 0.67 \\
$dropout=0.4$ & 6.14 & 4.82 & 0.52 & 0.59 \\
$dropout=0.5$ & 8.19 & 6.51 & 0.46 & 0.54 \\
$D_f=4$ & 2.62 & 2.12 & 0.76 & 0.82 \\
$D_f=8$ & 2.43 & 2.30 & 0.75 & 0.82 \\
$D_f=32$ & 2.60 & 1.99 & 0.65 & 0.70 \\
$D_f=64$ & 3.24 & 2.80 & 0.57 & 0.58 \\
\midrule
Ours & \textbf{2.37} & \textbf{1.48} & \textbf{0.77} & \textbf{0.83} \\
\bottomrule
\end{tabular}
\end{center}

\end{table}

\begin{figure}[t]
\begin{center}
   \includegraphics[width=0.85\linewidth]{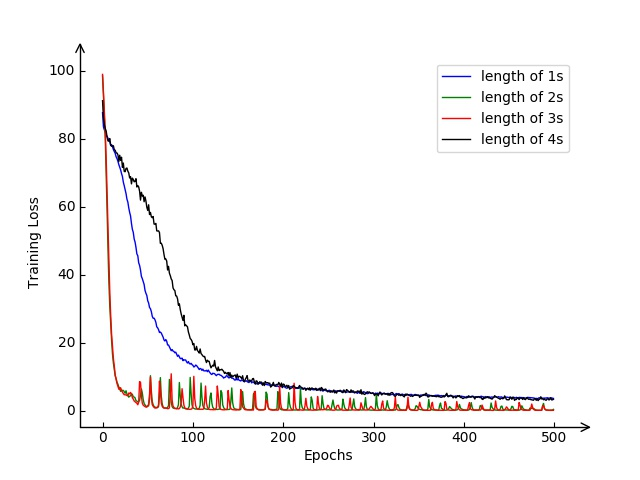}
\end{center}
   \caption{
  Losses of our model when training different length of pose fragments. From left to right are different length pose fragments from 1s to 4s (The x-axis is the training epochs and the y-axis is the loss).}
\label{fig:dif-length}
\end{figure}

\subsection{Performance Analysis}

\subsubsection{Matching Phase }
The training and validation Euclidean distance of Eqn.~\eqref{eq:loss} can be found in~\tabref{tab:hyper}. The matching system performs best with the combination of $(lr=0.0001,dropout=0.1, D_f=16)$, which achieves the minimal loss in the training set and validation set simultaneously. To test the ability to find the consistency of the music and pose clips, we measure the similarity of two features by the Euclidean distance between them.
To evaluate the matching accuracy from another perspective, we thus define the correlation accuracy as: i) the music and pose clip are corresponding if the distance between two features is less than $1$, ii) the music and pose clip are non-corresponded if the distance is larger than $1$.

\begin{figure}[t]
\begin{center}
   \includegraphics[width=0.9\linewidth]{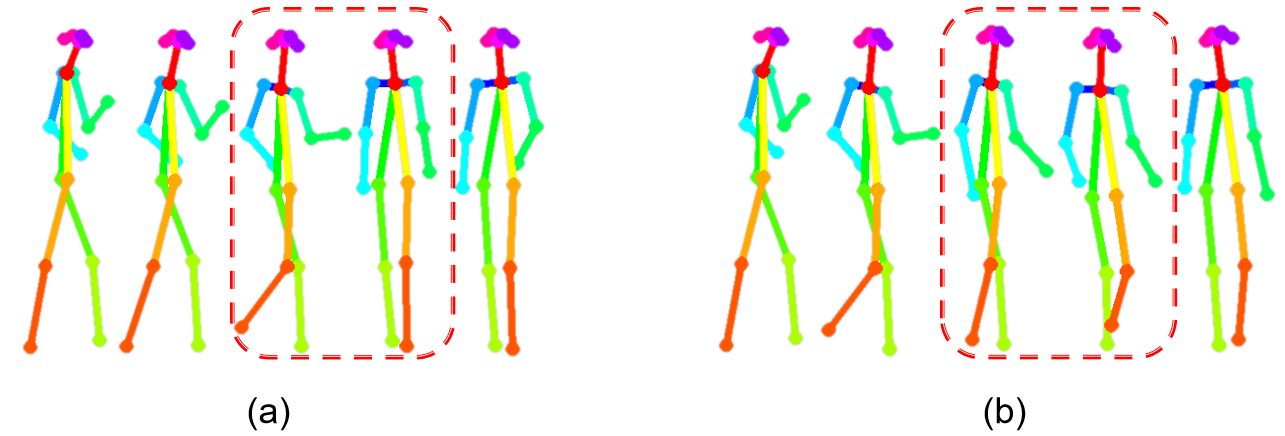}
\end{center}
   \caption{The pose sequence after spatial alignment: (a) the discontinuity (in the red dotted box) between two adjacent pose fragments. (b) the optimized pose sequence.}
\label{fig:spa}
\end{figure}

\begin{figure}[t]
\begin{center}
   \includegraphics[width=1\linewidth]{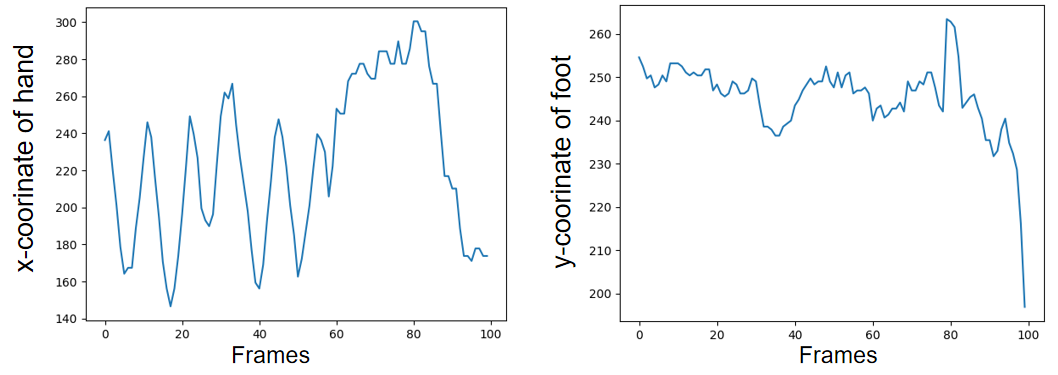}
\end{center}
   \caption{
   The movement tendency of the dance. The top is the x-axis trend of the hand node and the bottom is the y-axis trend of the foot node. The dance movements have strong periodicity in some periods (top) and not in others (bottom).}
\label{fig:tendency}
\end{figure}

The length of pose fragment is a key factor that affects the matching and generation quality.
To test the feasibility of our framework, we experiment with different lengths of $\textbf{x}^s$ from $1s$ to $4s$. It is obvious that all the lengths of pose fragments can converge in a short period of training time (see \figref{fig:dif-length} for details). And the matching model with a short length of $\textbf{x}^s$ shows a more rapid converge than that of a large length.
However, there exists a notable trade-off between the matching phase and the generation phase.
In the spatial-temporal alignment module $\Phi_{st}$ of generation stage, we notice that the results $\hat{\textbf{p}}^t$ are sometimes unstable when the length of $\textbf{x}^s$ is extremely short (\eg, $1$s) and unable to achieve a complex movement,~\eg, turning the body around. Thus to make the system consistent, we choose the length of pose fragment with more than $1$s.

\begin{figure}[t]
\begin{center}
   \includegraphics[width=0.9\linewidth]{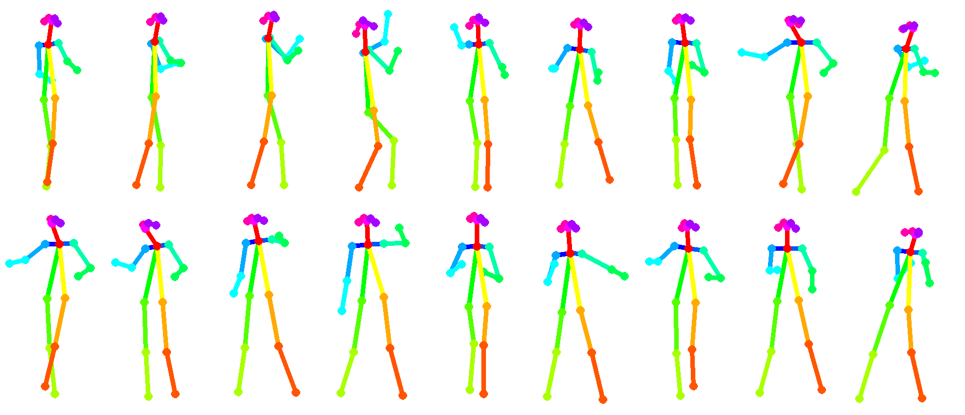}
\end{center}
   \caption{The predicted pose sequence of different music: top sequences matched to the music \emph{Updown Funk}, bottom sequences matched to the music \emph{Follow Me}. Our model has the potential to predict complex pose movement.}
\label{fig:dif-music}
\end{figure}

The other intuitive way to evaluate the matching results is the exhibition of skeleton movements.
After the model $\Phi_{st}$ is trained, we experiment with different kinds of music $\textbf{a}^t$ to the system and exhibit two typical $\textbf{p}^t$ at the same moment (see \figref{fig:dif-music}). We observe that various $\textbf{p}^t$ in different music closely matching the content of the music and the skeletons act in a steady motion without abrupt changes.
Remarkably, our model has the potential to predict complex pose movements, which is extremely lacking in the end-to-end learning scheme. For example, there exists a backside action (top of \figref{fig:dif-music}) in the generated results.

\begin{figure}[t]
\begin{center}
   \includegraphics[width=1.0\linewidth]{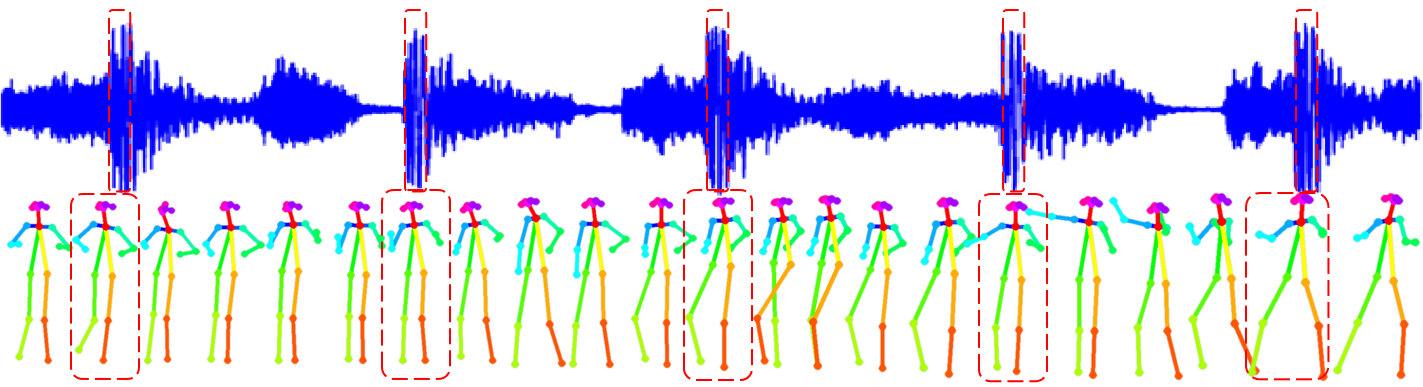}
\end{center}
   \caption{The pose sequence after temporal alignment: matched dance movements at different musical moments. After the processing of the temporal algorithm, the rhythms (in the red dotted box) of the pose sequence and music are aligned.}
\label{fig:beat_case}
\end{figure}

\subsubsection{Generation Phase}
In the generation phase of pose sequence $\textbf{p}^t$, we need to retrieve the adequate pose fragment for database $\mathbb{D}$. However, these fragments are not on the same scale owing that they are collected from different video frames. To solve this problem, we propose a normalization algorithm to unify the pose fragment into the same scale. For each selected pose fragment, we record the maximum distance between the nose keypoint and the feet keypoint. Then we resize the retrieved fragments as the same size as the target person. The ratios of different pose fragments are then used to expand the frame equally in the $x$ and $y$ axis.

\textbf{Spatial alignment.} We therefore visualize the spatial misalignment in \figref{fig:spa} (a). It is obvious that there exists an abnormal movement in the retrieved pose sequence. As stated in the spatial alignment module, we extract two dance movements of hand keypoint and foot keypoint in \figref{fig:tendency} respectively. It can be concluded that the hand keypoint moves periodically and the foot keypoint moves nonperiodic. Based on this finding, we conduct the proposed spatial alignment with the trend term $M_t$ and periodic term $S_t$ of $i$th frame. The refined results with our spatial alignment module are shown in \figref{fig:spa} (b), in which the human pose moves more stably and turns around in a more reasonable way.

\textbf{Temporal alignment.} To further align musical beats with pose movements, we use the temporal alignment for consistency, which is shown in~\figref{fig:beat_case}. The processed music can be viewed in blue, in which the beats are viewed in the red dotted box. To align the pose sequences with musical beats, the red dotted box shows the most significant movement in the local window of $\omega_c$ frames.
To be specific, we set the parameter $\omega_c$ as the interval length of the adjacent musical beats, which remains constant in each piece of music.
In this manner, it can be concluded that the changing of movement are naturally aligned with the musical beats and meanwhile overall pose moving trends are greatly maintained.
For example, at the first beat, the right foot of the skeleton moves a large step simultaneously.

\begin{figure}[t]
\begin{center}
   \includegraphics[width=1.0\linewidth]{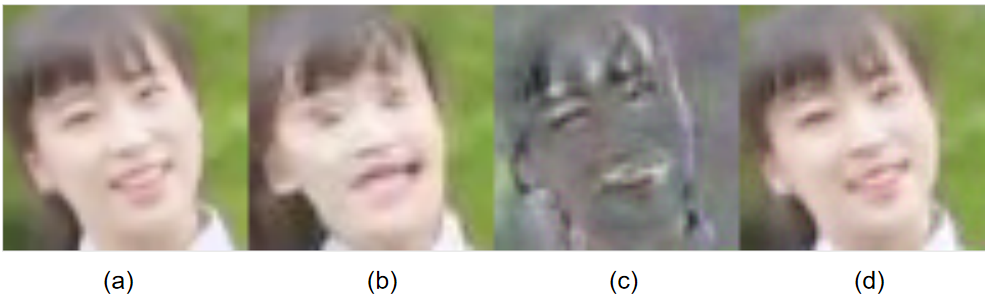}
\end{center}
   \caption{Visualized results with FaceGAN: (a) the groundtruth image. (b) the generated image. (c) the gradient generated by the FaceGAN generator. (d) the image optimized by FaceGAN. }
\label{fig:facegan}
\end{figure}

\begin{figure}[t]
\begin{center}
   \includegraphics[width=0.9\linewidth]{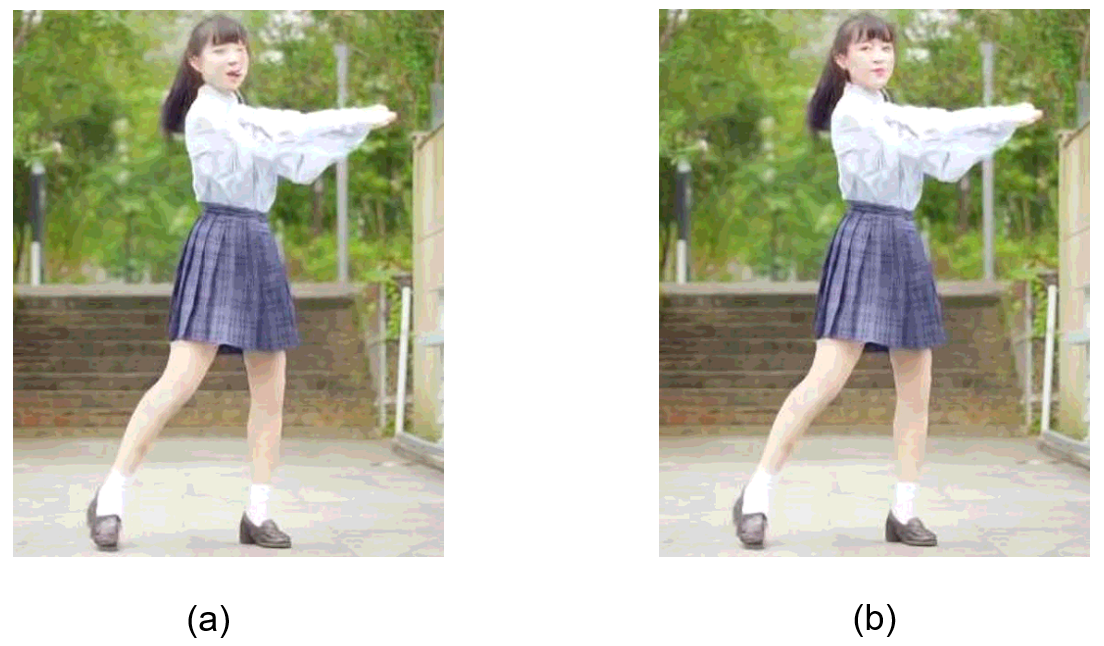}
\end{center}
   \caption{The example of synthesizing target dance image: (a) the image of the target character generated from the pose. (b) the generated image with the face region enhanced by FaceGAN.}
\label{fig:synthesis}
\end{figure}

\begin{figure}[t]
\begin{center}
   \includegraphics[width=1.0\linewidth]{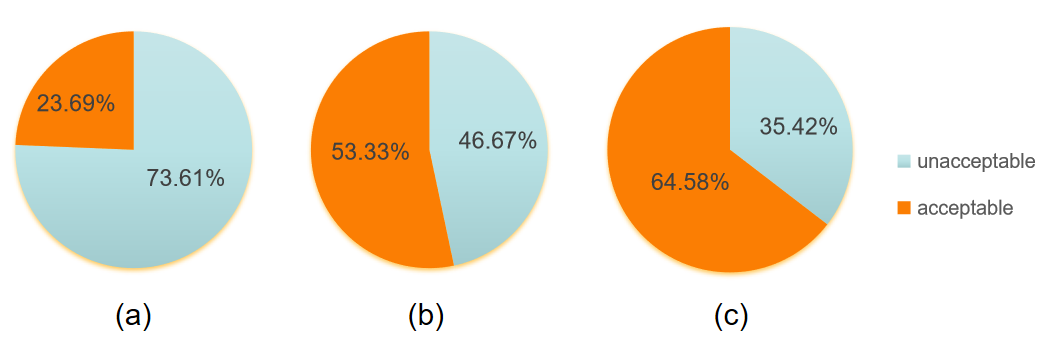}
\end{center}
   \caption{User study. (a) The voting result of the dancing videos obtained without any optimization. (b) The voting result of the dancing videos obtained by linear interpolation. (c) The voting result of the dancing videos obtained after the processing of the spatial-temporal alignment module.}
\label{fig:ablation}
\end{figure}

\textbf{Video imagination.}
After getting the final pose sequence $\hat{\textbf{p}}^t$, we synthesize the target person dancing video $\textbf{v}^t$ with the resolution of $512\times512$ in the imagination module, as shown in \figref{fig:synthesis} (a). It can be seen that our model can be generalized the complex scenarios with higher light intensities. However, the face regions are usually blurry owing that our human beings pay much more attention to the face regions than other regions. Thus we adopt the FaceGAN in~\figref{fig:facegan} to adopt the gradient map of the face region as input. With this local enhancement, the local regions are further concatenated with the holistic image, which is viewed in \figref{fig:synthesis} (b)).

\begin{figure*}[t]
\begin{center}
   \includegraphics[width=1\linewidth]{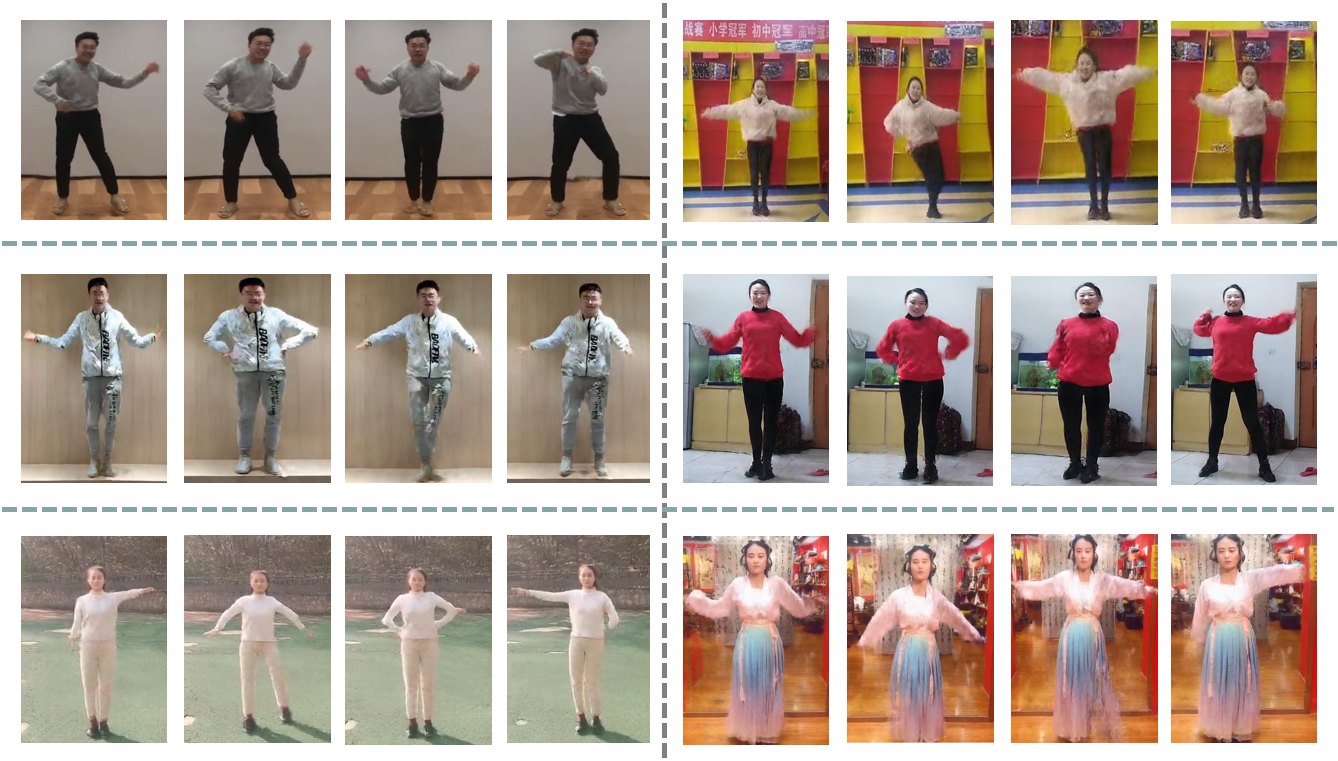}
\end{center}
   \caption{Synthesized results of multiple targets: Images of different human dances generated in different music. By inputting different music and videos of different persons, our proposed approach is able to generate satisfactory results that match human subjective evaluations.}
\label{fig:fin-result}
\end{figure*}



\subsection{Time efficiency}
 Our system consists of the cross-modal alignment, spatial-temporal alignment and imagination module. With the resolution of $512\times 512$ images, the computation costs of our two alignment modules are $23.59$GFLops and the cost of the imagination module is $373.28$GFLops. In the process of generating dancing sequences, our proposed model generates $48.8$ frames per second, which is relatively light-weight. In the process of final video synthesis, our proposed imagination module generates $8.59$ frames per second.

\subsection{Subjective Evaluation}

To evaluate the quality of the generated dancing videos, we conduct a user study to measure the effectiveness of our proposed spatial-temporal alignment module. We evaluate two different versions of our method: 1) pose sequences without any optimization, 2) pose sequences with linear fitting interpolation. We develop a simple user interface to play these two videos with the same musical pieces and choose the option and the subsequent studies are conducted in the same setting.

To make a fair comparison, we randomly sample 20 video clips from the generated dancing videos of each version. Each participant watches the video clips and chooses one of the two options: 1) the synthesized dancing video is realistic, 2) the synthesized dancing video is unrealistic.
As shown in \figref{fig:ablation} (a), without any optimization, $26.39\%$ of the votes indicate the synthesized dancing videos are realistic, while $73.61\%$ of them are unrealistic. The reason for these voting results is that the dance movement shows abnormal changes, which are unnatural to human subjective evaluations.

The new voting results obtained after the post-processing of the pose sequence with linear fitting interpolation are shown in \figref{fig:ablation} (b). $53.33\%$ of participants vote that the synthesized dancing videos are realistic, while $46.67\%$ do not.
After performing the proposed alignment module, this new voting result obtained after spatial alignment is shown in \figref{fig:ablation} c). $64.58\%$ of participants vote that the synthesized dancing videos are realistic, while $35.42\%$ participants do not.




\begin{figure}[t]
\begin{center}
   \includegraphics[width=1\linewidth]{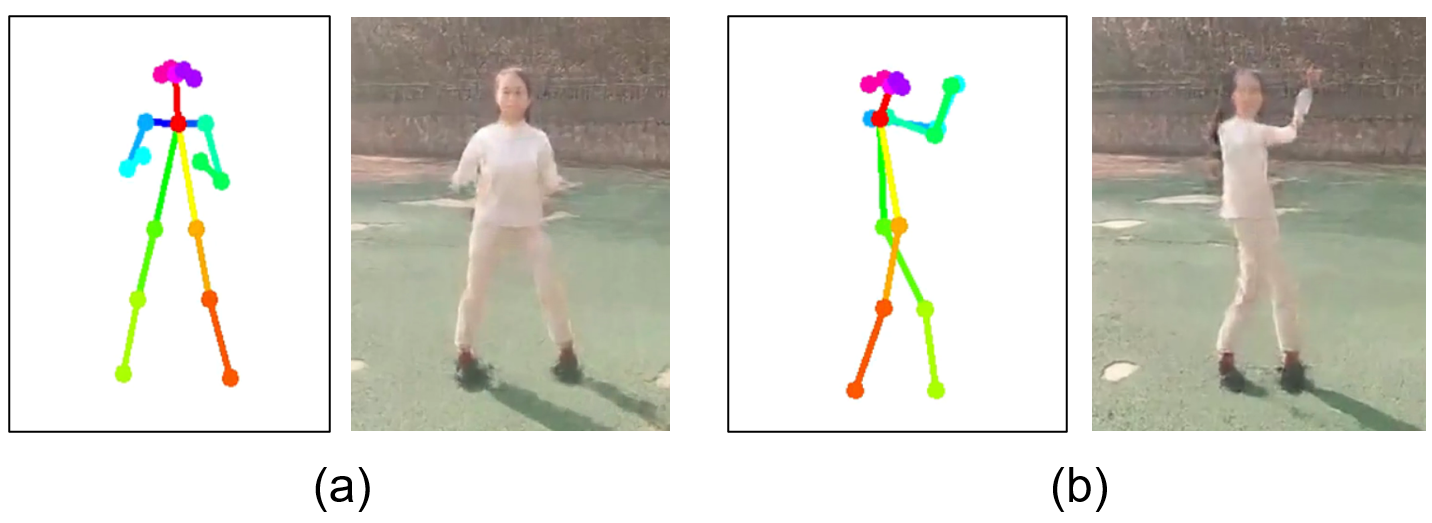}
\end{center}
   \caption{The failure cases of synthesizing images: (a) The arms are blurry. (b) The face and arms are unnatural.}
\label{fig:synfail}
\end{figure}

\subsection{Generalizations and Failure Modes}

\textbf{Video syntheses under different scenarios.}
Benefiting from the alignment modules and the pose normalization, our approach is able to provide synthesized videos in various scenarios. As shown in~\figref{fig:fin-result}, the generated videos under different conditions show natural movements and clear dancing poses, which answers the question in~\secref{sec:intro}: our proposed method can generate reliable imaginations with the input of music pieces. For example, with the confusing background to the human bodies, our approaches generate stable pose sequences while keeping the background relatively static, especially in the last group of special costumes.

\textbf{Failure modes.} Although our proposed method generates reasonable dancing videos, there are still some limitations in our system. As shown in~\figref{fig:synfail} a), the generated arms are blurry in some complex scenarios, which forces us to improve the quality of the generator. However, our two alignment modules provide satisfactory results of dancing poses, which could be reliable guidance for the imagination module.
In the second case of ~\figref{fig:synfail} b), when performing the complicated poses, the imagination still faces challenges in distinguishing the overlapping parts with similar visual appearances, which is left for our future work.

\section{Conclusions}\label{sec:conclusion}
In this paper, we focus on the audio-inspired video synthesis, which aims to generate reliable dancing videos corresponding to the input audio. To solve this important problem, we first collect a data set with musical audios and video clips, while combing the pose sequences at the same time.
Based on this dataset, we make an attempt to align these two media with two alignment modules,~\ie, the cross-modal alignment module and the spatial-temporal alignment module. Considering the audio and video frames are two completely different modalities, we process these two media with different encoders and take the pose as an intermediary to build the connections. Thus the relationship between the audio and pose sequences is established. Then this learned relationship helps the retrieval of pose fragments with given test audio, and we concatenate these fragments as the dancing guidance. However, there exists strong misalignment between the pose fragments and audio beats, we thus adopt the spatial-temporal alignment module to align the pose sequences with musical beats and changing steadily in a more natural manner. With the generated pose sequences as guidance, we finally generate the imagined dancers moving vividly aligning with the input music. Experimental evidence demonstrates that our method shows a reliable matching correlation of different types of media and generates realistic videos that match the human subjective evaluations.

\ifCLASSOPTIONcaptionsoff
  \newpage
\fi

\section*{Acknowledgment}
We would like to express our sincere appreciation to Tengfei Shi, Weili Lu, Jing Guo and Zhengdong Zhang for their volunteer work in providing the dancing videos used in this research.
This work was supported by grants from National Natural Science Foundation of China (No.61922006) and Baidu academic collaboration program.



%
\ifCLASSOPTIONcaptionsoff
  \newpage
\fi

\bibliographystyle{IEEEtran}
\bibliography{egbib}

%

\begin{IEEEbiography}[{\includegraphics[width=1in,height=1.25in,clip,keepaspectratio]{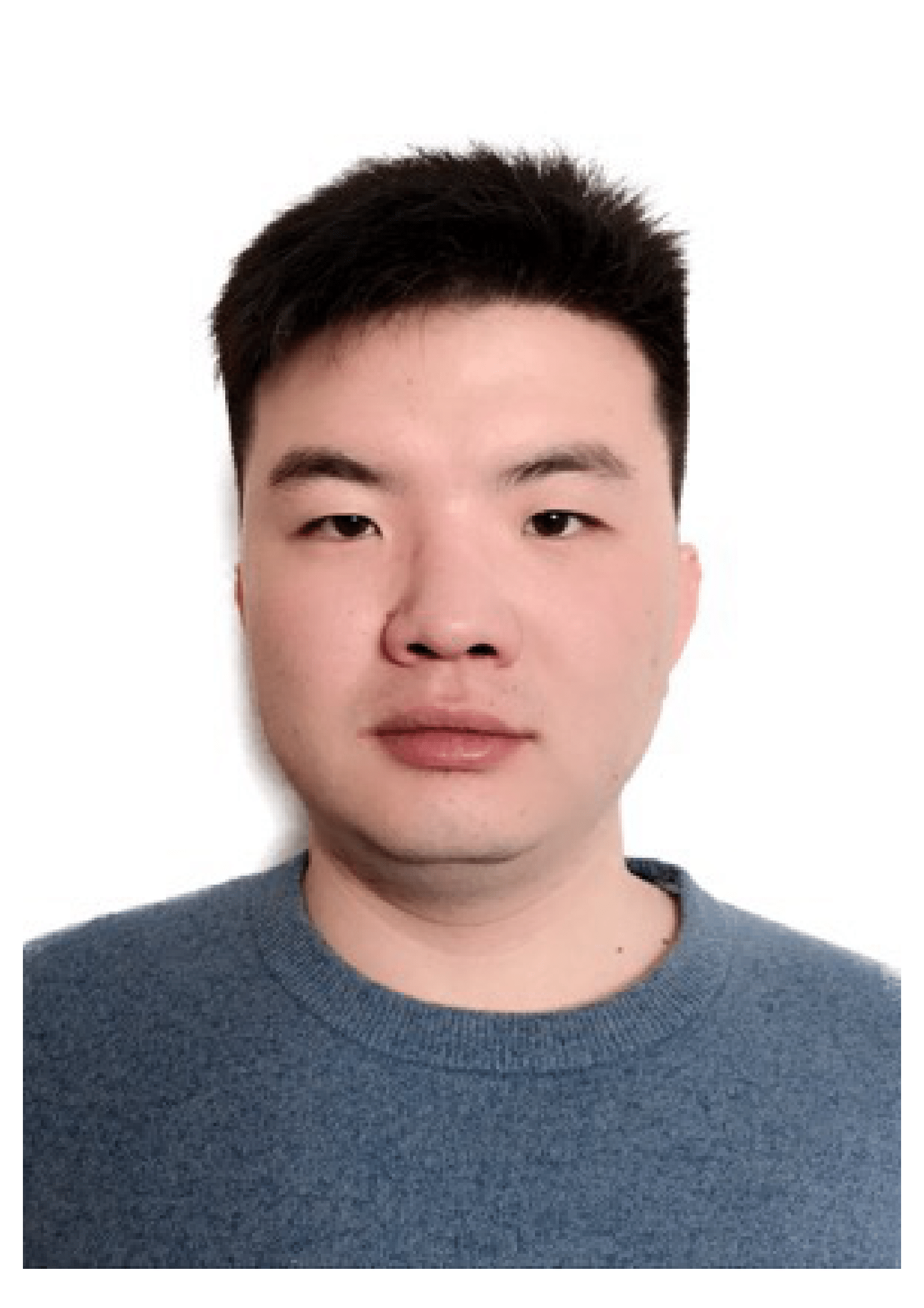}}]{Xin Guo}
is currently pursuing the master degree with the State Key Laboratory of Virtual Reality Technology and System, School of Computer Science and Engineering, Beihang University. His research interests include computer vision and cross-modal learning.
\end{IEEEbiography}

\begin{IEEEbiography}[{\includegraphics[width=1in,height=1.25in,clip,keepaspectratio]{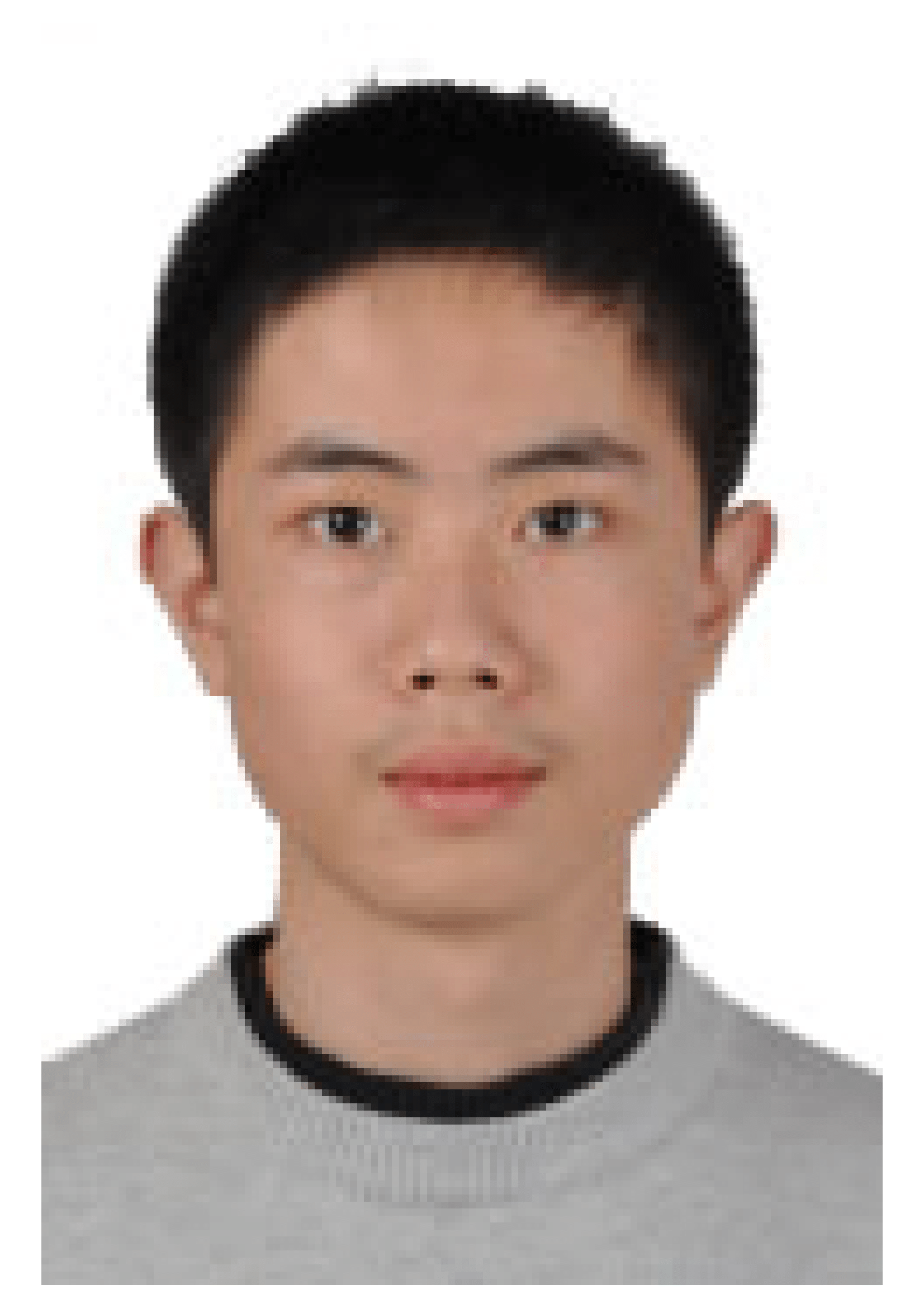}}]{Yifan Zhao}
is currently pursuing the Ph.D. degree with the State Key Laboratory of Virtual Reality Technology and System, School of Computer
Science and Engineering, Beihang University. He received the B.E. degree from Harbin Institute of Technology in Jul. 2016. His research interests include computer vision and image understanding.
\end{IEEEbiography}

\begin{IEEEbiography}[{\includegraphics[width=1in,height=1.25in,clip,keepaspectratio]{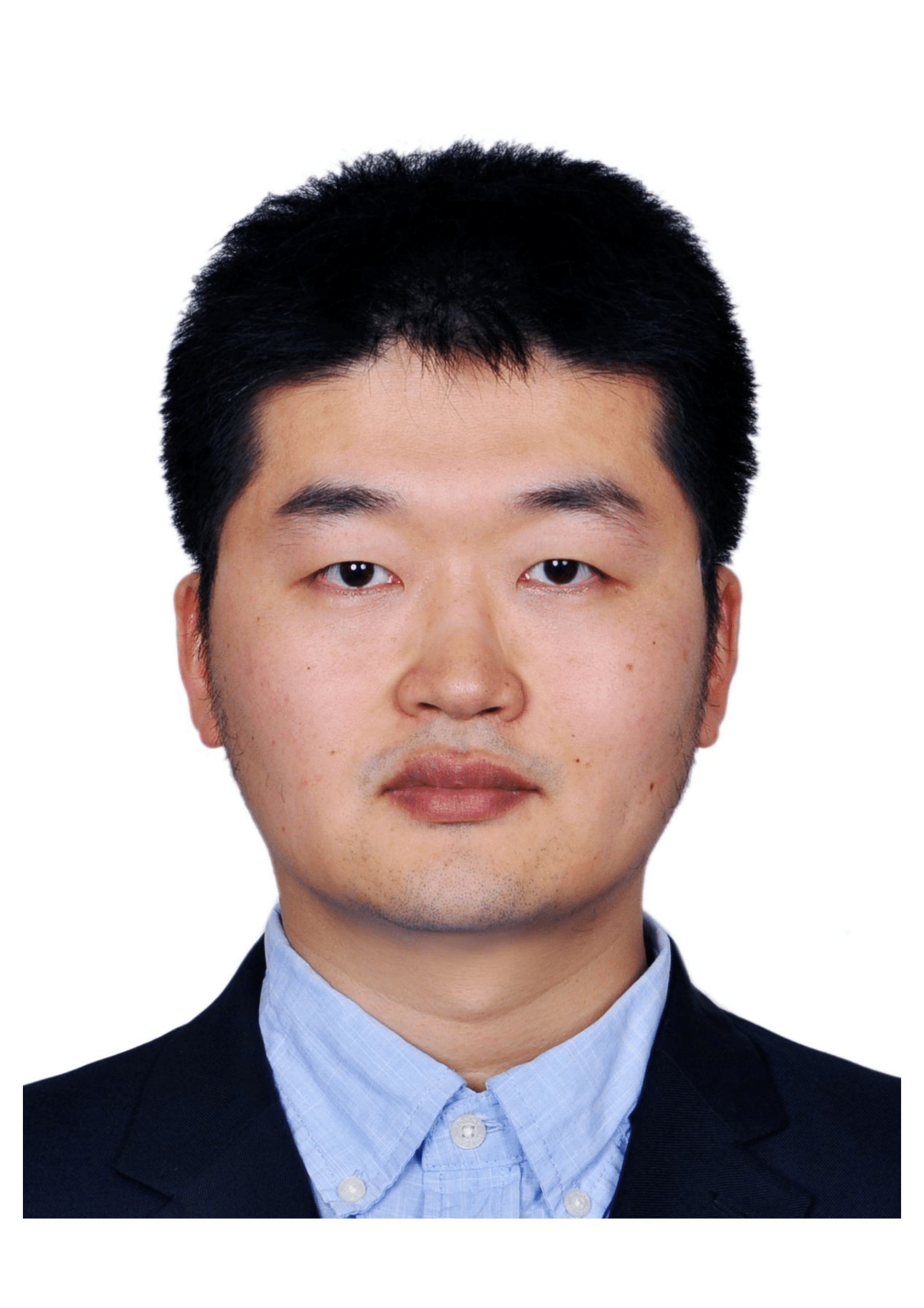}}]{Jia Li}
(M'12-SM'15) received the B.E. degree from Tsinghua University in 2005 and the Ph.D. degree from the Institute of Computing Technology, Chinese Academy of Sciences, in 2011. He is currently a Full Professor with the School of Computer Science and Engineering, Beihang University, Beijing, China. Before he joined Beihang University in Jun. 2014, he used to conduct research in Nanyang Technological University, Peking University and Shanda Innovations. He is the author or coauthor of over 70 technical articles in refereed journals and conferences such as TPAMI, IJCV, TIP, CVPR and ICCV. His research interests include computer vision and multimedia big data, especially the understanding and generation of visual contents. He is supported by the Research Funds for Excellent Young Researchers from National Nature Science Foundation of China since 2019. He was also selected into the Beijing Nova Program (2017) and ever received the Second-grade Science Award of Chinese Institute of Electronics (2018), two Excellent Doctoral Thesis Award from Chinese Academy of Sciences (2012) and the Beijing Municipal Education Commission (2012), and the First-Grade Science-Technology Progress Award from Ministry of Education, China (2010). He is a senior member of IEEE, CIE and CCF. More information can be found at http://cvteam.net.

\end{IEEEbiography}





\end{document}